\documentclass[journal]{IEEEtran}
\ifCLASSINFOpdf
\else
\fi

\usepackage{subfigure,multirow,array,graphicx,amssymb,amsmath,booktabs}
\usepackage{xcolor}
\usepackage[flushleft]{threeparttable}

\newcommand{\highlight}[1]{\textcolor{black}{#1}}
\newcommand{\Highlight}[1]{\textcolor{black}{#1}}
\newcommand{\hhighlight}[1]{\textcolor{black}{#1}}

\newcommand{\etal}{\textit{et al.}}

\begin{document}
%
\title{Facial Action Unit Detection via Adaptive Attention and Relation}

%
%
%

\author{Zhiwen~Shao,
        Yong~Zhou,
        Jianfei~Cai,~\IEEEmembership{Fellow,~IEEE,}
        Hancheng~Zhu,
        and~Rui~Yao
\thanks{Manuscript received February, 2022. This work was supported in part by the National Natural Science Foundation of China under Grant 62106268, in part by the High-Level Talent Program for Innovation and Entrepreneurship (ShuangChuang Doctor) of Jiangsu Province under Grant JSSCBS20211220, and in part by the Talent Program for Deputy General Manager of Science and Technology of Jiangsu Province under Grant FZ20220440. It was also supported in part by the National Natural Science Foundation of China under Grants 62272461, 62101555, and 62172417, in part by the Natural Science Foundation of Jiangsu Province under Grants BK20201346 and BK20210488, in part by the Shanghai Sailing Program under Grant 23YF1410500, and in part by the Fundamental Research Funds for the Central Universities under Grant 2021QN1072. The authors gratefully acknowledge the support of K.C.Wong Education Foundation, Hong Kong. (Corresponding author: Yong~Zhou.)}
\thanks{Z. Shao is with the School of Computer Science and Technology, China University of Mining and Technology, Xuzhou 221116, China, also with the Engineering Research Center of Mine Digitization, Ministry of Education of the People’s Republic of China, Xuzhou 221116, China, and also with the Department of Computer Science and Engineering, Shanghai Jiao Tong University, Shanghai 200240, China (e-mail: zhiwen\_shao@cumt.edu.cn).
}
\thanks{Y. Zhou, H. Zhu, and R. Yao are with the School of Computer Science and Technology, China University of Mining and Technology, Xuzhou 221116, China, and also with the Engineering Research Center of Mine Digitization, Ministry of Education of the People’s Republic of China, Xuzhou 221116, China (e-mail: \{yzhou; zhuhancheng; ruiyao\}@cumt.edu.cn).
}
\thanks{J. Cai is with the Faculty of Information Technology, Monash University, Victoria 3800, Australia (e-mail: jianfei.cai@monash.edu).}
}

%
%

\markboth{IEEE Transactions on Image Processing,~Vol.~X, No.~X, X}%
{Shell \MakeLowercase{\textit{et al.}}: Bare Demo of IEEEtran.cls for IEEE Journals}
%



\maketitle

\begin{abstract}

Facial action unit (AU) detection is challenging due to the difficulty in capturing correlated information from subtle and dynamic AUs. Existing methods \Highlight{often} resort to the localization of correlated regions of AUs, \Highlight{in which predefining} local AU attentions by correlated facial landmarks often discards \Highlight{essential parts}, or \Highlight{learning} global attention maps often contains irrelevant \Highlight{areas. Furthermore, existing} relational reasoning methods often employ common patterns for all AUs while ignoring the specific \Highlight{way} of each AU.
To tackle these limitations, we propose a novel adaptive attention and relation (AAR) framework for facial AU detection. Specifically, 
we propose an adaptive attention regression network to regress the global attention map of each AU under the constraint of attention predefinition and the guidance of AU detection, 
which is beneficial for capturing both \Highlight{specified dependencies by landmarks in} strongly correlated regions and \Highlight{facial globally distributed dependencies in} weakly correlated regions. Moreover, considering the diversity and dynamics of AUs, we propose an adaptive spatio-temporal graph convolutional network to simultaneously reason the independent pattern of each AU, the inter-dependencies among AUs, as well as the temporal dependencies. Extensive experiments show that our approach (i) \highlight{achieves competitive performance}
on challenging benchmarks 
including BP4D, DISFA, and GFT in constrained scenarios and Aff-Wild2 in unconstrained scenarios,
and (ii) can precisely learn the regional correlation distribution of each AU.

\end{abstract}


%
\IEEEpeerreviewmaketitle

\section{Introduction}
%
%
%
%
\IEEEPARstart{F}{acial} action unit (AU) detection has recently attracted increasing attention in the communities of computer vision and affective computing~\cite{li2018eac,shao2018deep,niu2019local,li2019semantic}, due to the extensive applications in relevant areas like medical care and digital entertainment. 
Each facial AU is associated with one or more local muscle actions, as defined in the facial action coding system (FACS)~\cite{ekman1997face}. The subtle appearance changes caused by AUs are different across facial identities and expressions. Besides, different AUs are often spatially co-occurred or mutually exclusive in certain expressions, and adjacent video frames often have similar AUs.
In literature AU detection has remained a challenging problem due to the difficulty of modeling the subtlety, diversity, and dynamics of AUs.

\begin{figure}
\centering\includegraphics[width=\linewidth]{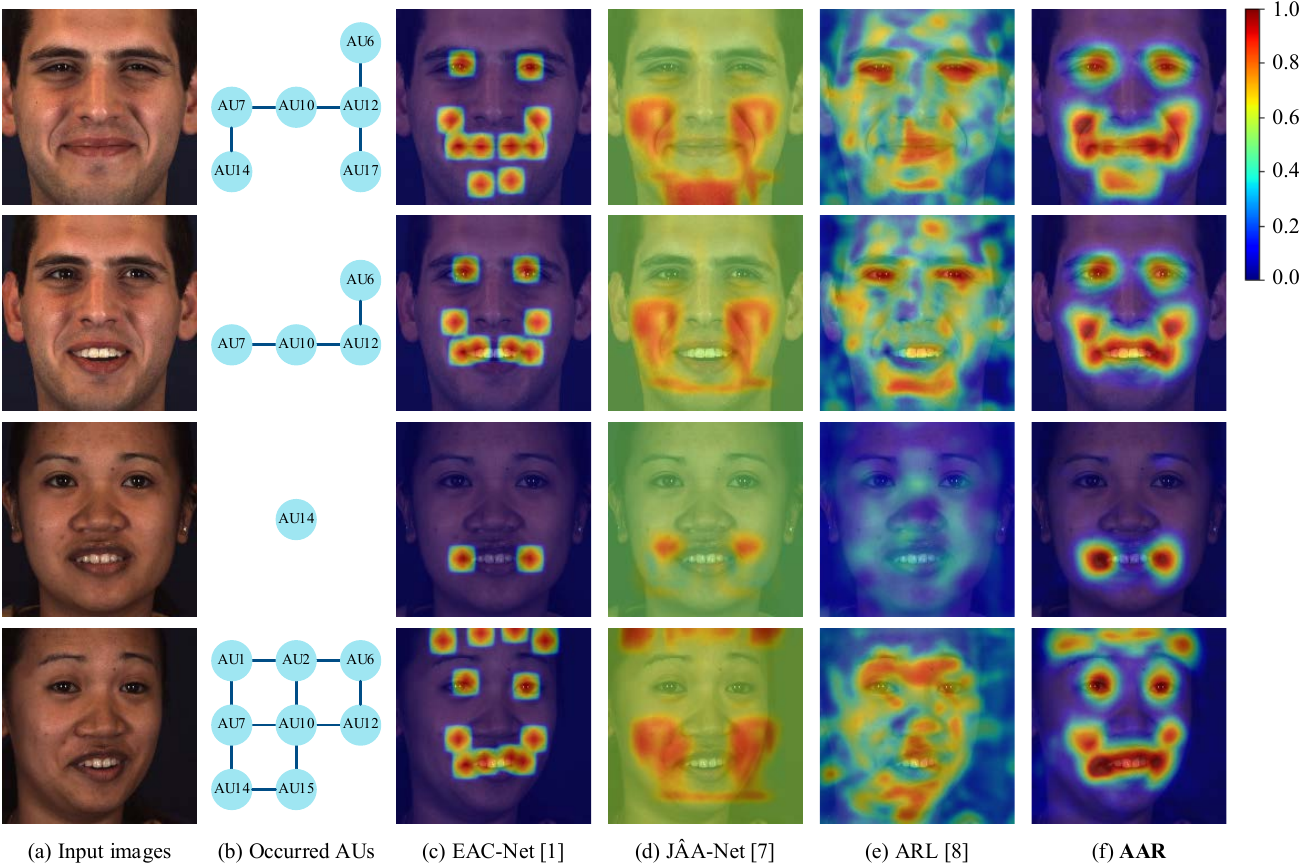}
\caption{Comparisons with state-of-the-art AU regional localization methods for several example BP4D~\cite{zhang2014bp4d} images with happy expression. (c)-(f) show the combined attention maps of occurred AUs, in which EAC-Net~\cite{li2018eac} uses predefined local attentions, JÂA-Net~\cite{shao2021jaa} refines the predefined attentions, ARL~\cite{shao2019facial} learns global attentions, and our AAR integrates the advantages of local attention predefinition and global attention learning. Attention weights are visualized using the colors shown in the color bar, which are overlaid on the input images for better viewing.}
\label{fig:motivation}
\end{figure}

Some AU detection works employ regional localization methods to capture the correlated regions of AUs. Li \etal~\cite{li2018eac,li2017action} predefined local region of interest (ROI) of each AU centered around correlated facial landmarks, in which the attention weight of the position farther away from the ROI center is lower, as shown in Fig.~\ref{fig:motivation}(c). Considering the ROI size and attention distribution should be varied across AUs, Shao \etal~\cite{shao2018deep,shao2021jaa} adaptively refined the predefined attention map of each AU during the training of AU detection. However, 
the refined attentions look like the smoothing of the predefined attentions, in which other regions far away from the predefined ROIs are treated with equal attention so that the included relevant regions are not highlighted. On the other hand, Shao \etal~\cite{shao2019facial} discarded the prior constraints of landmarks, and adaptively learned global attention map of each AU with the supervision of AU detection. Although the learned attention map can capture the correlated regions in the global face, some irrelevant regions are also included.    

Another line of AU detection is to model the correlations among AUs. Earlier works~\cite{tong2008learning,li2013simultaneous,wang2013capturing,wang2013capturing2} employ probabilistic graphical models (PGMs)~\cite{pearl1988probabilistic} including Bayesian network (BN), dynamic Bayesian network (DBN) and restricted Boltzmann machine (RBM) to model local pairwise AU dependencies or global AU dependencies. These methods resort to hand-crafted features, and often only focus on local pairwise relations or implicit global relations, which limit the performance of relational reasoning. Recently, by exploiting the prevailing deep learning technology, Corneanu \etal~\cite{corneanu2018deep} proposed a deep structure inference network (DSIN) to infer the structure among AUs by iterative message passing. Li \etal~\cite{li2019semantic} predefined AU relation graph based on statistical AU dependencies in AU datasets, and Song \etal~\cite{song2021uncertain} adaptively learned a probabilistic mask to capture AU correlations. However, these methods adopt shared parameters among all AUs during relational reasoning, which neglects the specificity and dynamics of each AU. For example, AU 14 occurs independently in a frame but co-occurs with other AUs in another frame for the woman in Fig.~\ref{fig:motivation}. 

To tackle the above limitations, we propose a novel \textbf{Adaptive Attention and Relation (AAR)} framework for AU detection. In particular, we propose an adaptive attention regression network to simultaneously regress the global attention map of each AU and predict the occurrence probability of each AU. Each attention map is encouraged to regress to the predefined attention map by landmarks, and is also adaptively learned with the guidance of AU detection. This is beneficial for precisely learning the regional correlation distribution, in which \Highlight{specified dependencies by landmarks in} strongly correlated regions and \Highlight{facial globally distributed dependencies in} weakly correlated regions are both captured.

Moreover, we propose an adaptive spatio-temporal graph convolutional network (GCN) for relational reasoning, in which the inputs are the feature of each AU learned by the adaptive attention regression network. During the spatio-temporal graph convolution, the adjacent relationship of the AU relation graph as well as a parameter pool are adaptively learned and are shared by all AUs to reason the inter-dependencies among AUs. By an adaptively learned disentanglement matrix, the parameters of each AU can be extracted from the parameter pool, so as to capture AU-specific patterns. Besides, gated recurrent unit (GRU)~\cite{cho2014properties} is combined with the graph convolution to reason the temporal dependencies. 

The main contributions of this paper are threefold:
\begin{itemize}
    \item We propose an adaptive attention regression network by integrating the advantages of local attention predefinition and global attention learning, which can capture both \Highlight{predefined dependencies by landmarks in} strongly correlated regions and \Highlight{facial globally distributed dependencies in} weakly correlated regions.
    \item We propose an adaptive spatio-temporal graph convolutional network to simultaneously reason the specific pattern of each AU, the inter-dependencies among AUs, as well as the temporal correlations. To our knowledge, AU-specific patterns reasoned in graph neural networks has not been explored before.
    \item Extensive experiments on benchmark datasets show that our approach \highlight{achieves comparable performance}
    in both constrained scenarios and unconstrained scenarios, and can accurately learn the regional correlation distribution of each AU.
\end{itemize} 

\begin{figure*}
\centering\includegraphics[width=\linewidth]{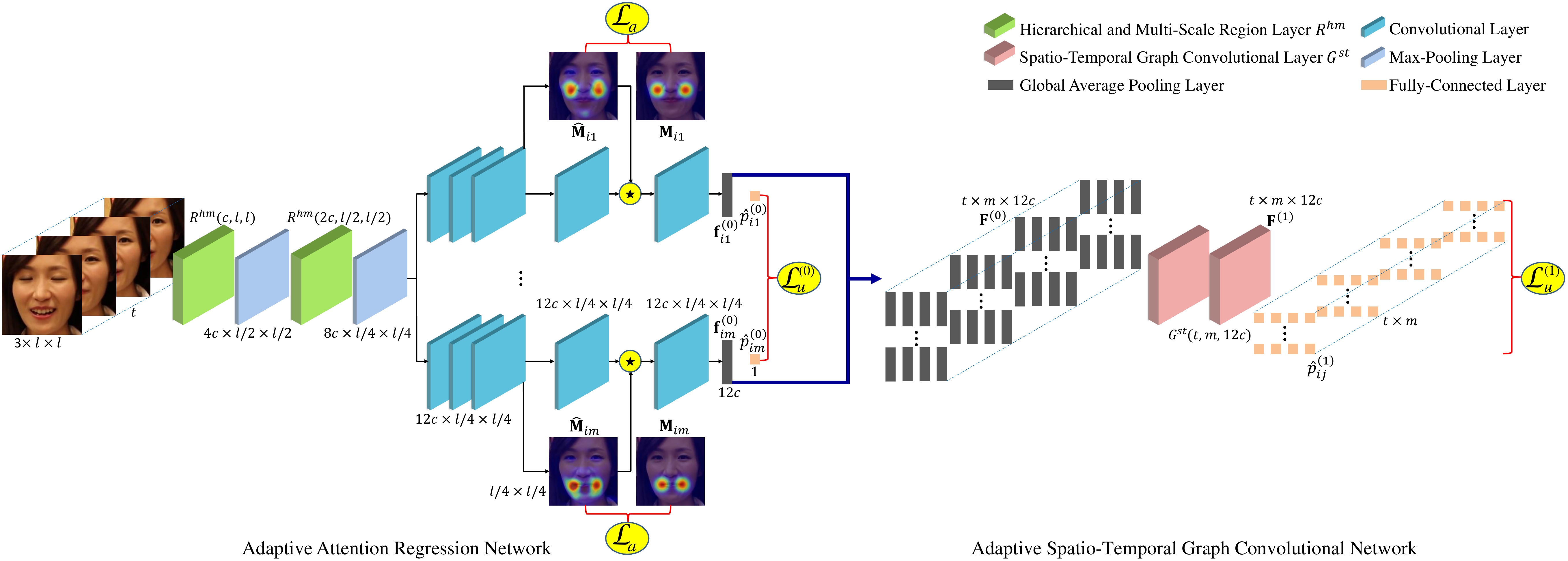}
\caption{The architecture of our AAR framework. Given a sequence of $t$ frames, our framework first extracts features $\mathbf{f}^{(0)}_{ij}$ of $m$ AUs for each frame by the proposed adaptive attention regression network, and further uses the proposed adaptive spatio-temporal graph convolutional network to simultaneously predict AU occurrence probabilities $\hat{p}_{ij}^{(1)}$ of all the $t$ frames. The adaptively regressed attention maps are overlaid on one input frame for better viewing, in which the shown examples $\widehat{\mathbf{M}}_{i1}$ and $\widehat{\mathbf{M}}_{im}$ correspond to AU 6 and AU 15, respectively. ``$\star$'' denotes the element-wise multiplication of the attention map and each feature map channel. The expression $c'\times l'\times l''$ denotes the dimensions of the corresponding layer are $c'$, $l'$, and $l''$, respectively.}
\label{fig:AAR_framework}
\end{figure*}

\section{Related Work}
\label{sec:relatedwork}

We review the previous techniques that are closely relevant to our work, including regional localization based AU detection and relational reasoning based AU detection.

\subsection{Regional Localization Based AU Detection}

Current AU detection methods often crop square regions or employ regional attentions to extract correlated features. Since facial landmarks can predefine the locations of AUs via prior knowledge, Zhao \etal~\cite{zhao2016joint} selected regions centered at certain landmarks to extract scale-invariant feature transform (SIFT)~\cite{lowe1999object} features for each AU. Instead of selecting predefined regions, Zhao \etal~\cite{zhao2016deep} uniformly partitioned a feature map into local regions, and applied independent convolutional filters to process each local region. Liu \etal~\cite{liu2020relation} utilized an independent convolutional network to extract a feature from each predefined image patch, while Li \etal~\cite{li2019semantic} first used a convolutional network over the input image to extract a global feature map and then processed the features cropped from the global feature map. Ma \etal~\cite{ma2019r} predefined the bounding box of each AU, and integrated the typical objection detection task into AU recognition. These methods treat each position in the cropped region with the same importance, which may discard significant information during the feature learning.

Inspired by the great success of attention mechanism in the computer vision field~\cite{kuen2016recurrent,cao2015look,pedersoli2017areas}, some works exploit the attentions to extract AU-related features. Li \etal~\cite{li2018eac,li2017action} applied a fixed size and a fixed attention distribution to the predefined ROI of each AU, in which the position with a farther distance to the ROI center has lower attention. Due to the availability of AU intensities, Sanchez \etal~\cite{sanchez2018joint} employed the AU intensity to determine the amplitude and size of a Gaussian distribution so as to generate an attention map for each AU. These methods neglect the diverse changes of AU appearances across persons and expressions. Recently, \Highlight{Wang \etal~\cite{wang2018recurrent} proposed a recurrent convolutional shape regression method to progressively predict facial landmark locations, in which facial shape increments in different steps are learned jointly. Robinson \etal~\cite{robinson2019laplace} regarded facial landmark detection as a heatmap regression problem, and proposed a Laplace KL-divergence loss to penalize scattered responses with low confidences.} Shao \etal~\cite{shao2018deep,shao2021jaa} adopted an adaptive attention learning module to refine the predefined spatial attention of each AU \Highlight{in a regression manner}, and Shao \etal~\cite{shao2019facial} further directly learned spatial and channel-wise attentions of AUs without the prior knowledge. The refined attentions learned in~\cite{shao2018deep,shao2021jaa} cannot highlight the potentially correlated regions far away from the predefined ROIs, and the attentions learned in~\cite{shao2019facial} also capture some irrelevant regions. In contrast, our proposed adaptive attention regression network can capture both strongly correlated regions around the predefined ROIs and weakly correlated regions scattered in the face.

\subsection{Relational Reasoning Based AU Detection}

Considering the inherent correlations among AUs, many works perform AU detection by exploiting the AU relations. Due to the capability of modeling complex data distributions, PGMs~\cite{pearl1988probabilistic} have been extensively used for AU detection. Tong \etal~\cite{tong2008learning} utilized a Bayesian network to model the local pairwise dependencies such as co-occurrence and mutual exclusion among AUs. Li \etal~\cite{li2013simultaneous} employed a dynamic Bayesian network for joint facial AU detection, expression recognition and landmark tracking, in which the local AU dependencies and local relations between AUs and landmarks are both reasoned. Wang \etal~\cite{wang2013capturing} proposed an interval temporal Bayesian network to capture complex spatio-temporal relations, in which the facial expression is treated as a complex activity containing overlapping or sequential primitive facial events. Instead of only modeling pairwise relations, Wang \etal~\cite{wang2013capturing2} applied a RBM to capture both local AU dependencies and implicit global AU correlations. All these methods are based on hand-crafted features, which limits their performance of learning AU relations.

Recently, there are some AU detection works combining relational learning with the prevailing deep learning technology. Chu \etal~\cite{chu2017learning} and He \etal~\cite{he2017multi} used deep convolutional networks to extract the facial feature of each frame and then used long short-term memory (LSTM) networks to model the temporal correlations within consecutive frames. Corneanu \etal~\cite{corneanu2018deep} proposed the DSIN to capture AU relations by explicitly passing information between AU predictions. Li \etal~\cite{li2019semantic} and Liu \etal~\cite{liu2020relation} predefined the AU relation graph based on prior knowledge, and then used graph neural networks to reason the correlations among AUs. 
Fan \etal~\cite{fan2021g2rl} employed a graph convolutional network to learn facial shape patterns and the inter-dependencies of facial landmarks, in which the geometric knowledge can facilitate AU recognition. Song \etal~\cite{song2021uncertain} proposed an uncertain graph convolution with a probabilistic mask to capture both the individual dependencies among AUs and the uncertainties. Besides, Song \etal~\cite{song2021hybrid} proposed a hybrid message passing neural network by using a performance-driven Monte Carlo Markov Chain sampling method to learn the AU relation graph, in which different types of messages are dynamically combined to provide complementary information. However, these methods employ shared parameters among all AUs for relational reasoning, which ignores the specificity and dynamics of individual AUs.

\begin{figure*}
\centering\includegraphics[width=0.8\linewidth]{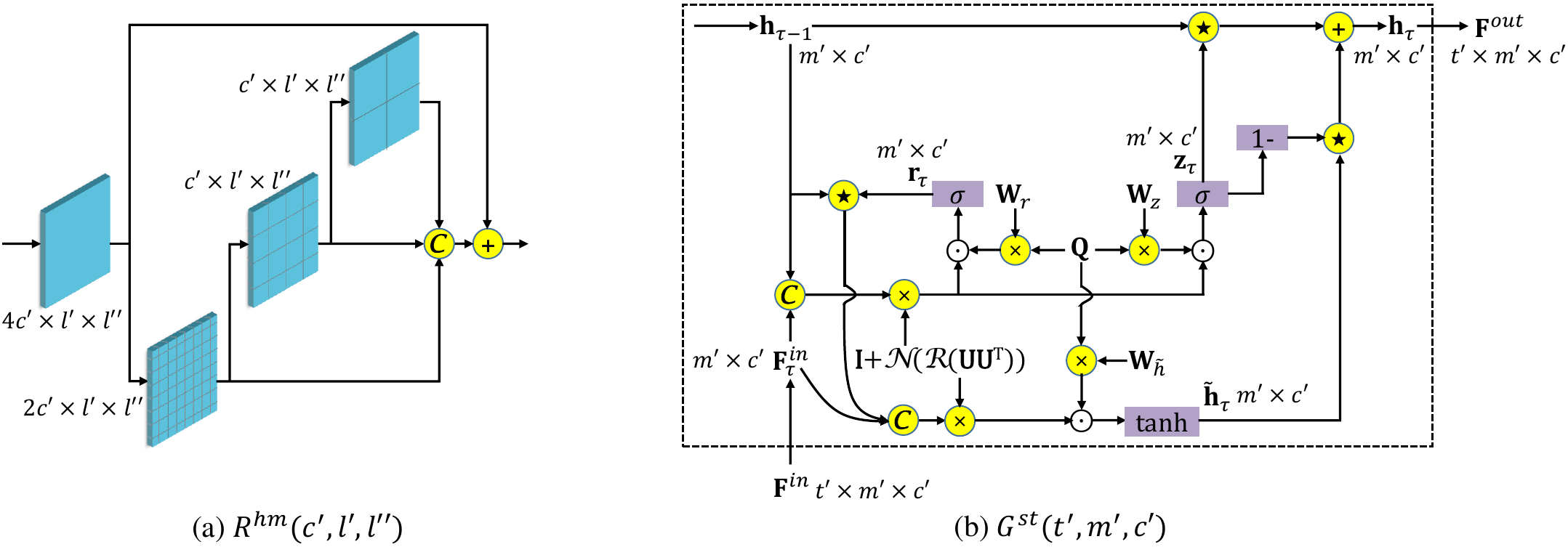}
\caption{The structures of a hierarchical and multi-scale region layer $R^{hm}(c',l',l'')$~\cite{shao2021jaa} and our proposed spatio-temporal graph convolutional layer $G^{st}(t',m',c')$. ``$\star$'' denotes the element-wise multiplication, ``$\mathcal{C}$'' denotes the concatenation, and ``$\sigma$'', ``$\tanh$'', and ``$1-$'' denote the Sigmoid function, the hyperbolic tangent function, and the operation subtracted by $1$, respectively.}
\label{fig:sub_blocks}
\end{figure*}

\section{AAR for Facial AU Detection}

\subsection{Overview}
Given consecutive $t$ frames 
with the same size of $3\times l\times l$, our main goal is to predict their AU occurrence probabilities $\{\hat{\mathbf{p}}_1^{(1)},\cdots,\hat{\mathbf{p}}_t^{(1)}\}$, where $\hat{\mathbf{p}}_i^{(1)}=(\hat{p}_{i1}^{(1)}, \cdots, \hat{p}_{im}^{(1)})$ for the $i$-th frame, and $m$ is the number of AUs. Fig.~\ref{fig:AAR_framework} illustrates the architecture of our framework. Specifically, we first propose an adaptive attention regression network to process the $t$ frames one by one, which simultaneously predict the global attention map $\mathbf{\widehat{M}}_{ij}$ and the occurrence probability $\hat{p}_{ij}^{(0)}$ for each AU. $\mathbf{\widehat{M}}_{ij}$ is encouraged to regress to the predefined attention map $\mathbf{M}_{ij}$, and is also adaptively learned under the supervision of the AU detection loss $\mathcal{L}_u^{(0)}$. By the adaptively regressed attention map $\mathbf{\widehat{M}}_{ij}$, we can precisely extract the feature $\mathbf{f}^{(0)}_{ij}$ of the $j$-th AU in the $i$-th frame. Then, we combine the features of all $m$ AUs of the $t$ frames as $\mathbf{F}^{(0)}$, and input it to our proposed adaptive spatio-temporal graph convolutional network. In each spatio-temporal graph convolutional layer $G^{st}$, the independent pattern of each AU, the inter-dependencies among AUs, as well as the temporal dependencies are all adaptively reasoned. With the AU detection loss $\mathcal{L}_u^{(1)}$, the outputs of this network are finally predicted AU occurrence probabilities.

\begin{figure}
\centering\includegraphics[width=\linewidth]{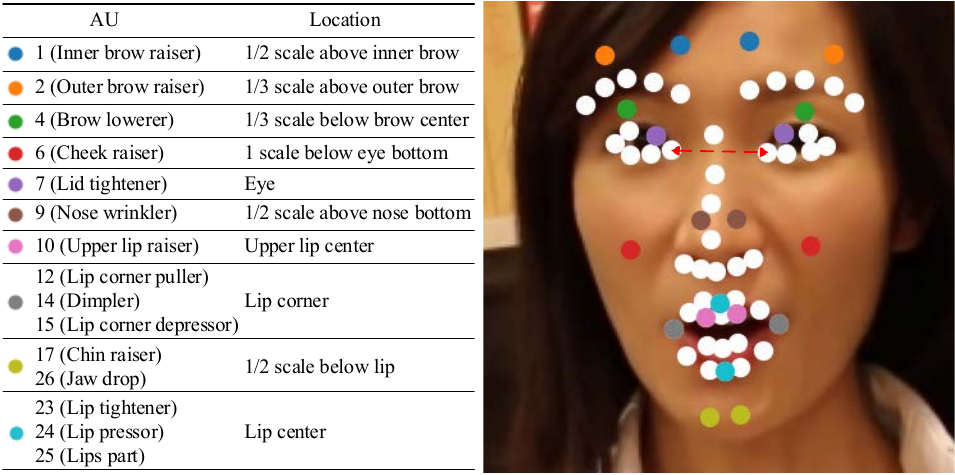}
\caption{Definition and visualization for the locations of AU centers, which are applicable to an aligned face whose eye centers are on the same horizontal line~\cite{li2018eac,shao2021jaa}. Each AU has two centers specified by two correlated facial landmarks. The red dotted line denotes ``scale'', which is the distance between two inner eye corners.}
\label{fig:au_landmark_location}
\end{figure}

\subsection{Adaptive Attention Regression Network}

As illustrated in Fig.~\ref{fig:AAR_framework}, one input frame first goes through two hierarchical and multi-scale region layers~\cite{shao2021jaa}, each of which is followed by a max-pooling layer. The detailed structure of a hierarchical and multi-scale region layer $R^{hm}$ is shown in Fig.~\ref{fig:sub_blocks}(a). It contains an input layer and three hierarchical intermediate layers with different scales of partitioned patches, in which each patch is processed with independent convolutional kernels to extract local features. The use of $R^{hm}$ is beneficial for extracting multi-scale features so as to adapt to various AUs in different local regions.

\subsubsection{Constraint of Attention Predefinition}

Similar to the structure of ARL~\cite{shao2019facial}, $m$ branches are then followed, each of which jointly performs attention map regression and AU prediction. Particularly, we adopt a one-channel convolutional layer following three convolutional layers to learn the attention map $\mathbf{\widehat{M}}_{ij}$ with size $l/4\times l/4$ for the $j$-th AU of the $i$-th frame, where $i=1,\cdots,t$, and $j=1,\cdots,m$. Considering correlated facial landmarks can precisely specify the locations of subtle AUs~\cite{li2018eac,shao2021jaa}, as shown in Fig.~\ref{fig:au_landmark_location}, we predefine the ground-truth attention map $\mathbf{M}_{ij}$ for each AU. Taking one AU center whose coordinate is $(\bar{a}_{ij}^{sub1},\bar{b}_{ij}^{sub1})$ as an example, we employ a Gaussian distribution centered on its location to generate the value at location $(a,b)$ of the attention map:
\begin{equation}
\label{eq:pre_att}
    M_{ijab}^{sub1} = \exp(-\frac{(a-\bar{a}_{ij}^{sub1})^2+(b-\bar{b}_{ij}^{sub1})^2}{2\delta^2}),
\end{equation}
where $\delta$ denotes the standard deviation, and $M_{ijab}^{sub1}\in (0,1]$. Then we incorporate the predefined attention maps of two AU centers by choosing the larger of the two attention weights at each location:
\begin{equation}
\label{eq:combine_pre_att}
    M_{ijab} = \max(M_{ijab}^{sub1}, M_{ijab}^{sub2}).
\end{equation}
Eqs.~\eqref{eq:pre_att} and~\eqref{eq:combine_pre_att} essentially assign a lower attention weight to a location farther away from both AU centers, as illustrated by the examples of $\mathbf{M}_{ij}$ in Fig.~\ref{fig:AAR_framework}.   

To preserve strongly correlated regions \Highlight{with pixel-level dependencies} specified by landmarks, we adopt an attention regression loss to encourage $\widehat{\mathbf{M}}_{ij}$ to close to $\mathbf{M}_{ij}$:
\begin{equation}
\label{eq:L_attention}
\mathcal{L}_{a} = \frac{1}{tm(l/4\times l/4)}\sum_{i=1}^t\sum_{j=1}^m\sum_{a=1}^{l/4}\sum_{b=1}^{l/4} (\widehat{M}_{ijab}-M_{ijab})^2.
\end{equation}

\subsubsection{Guidance of AU Detection}

We also expect weakly correlated regions distributed globally in the face, which are changed diversely across persons and expressions, can be adaptively captured by $\mathbf{\widehat{M}}_{ij}$. We element-wise multiply $\mathbf{\widehat{M}}_{ij}$ with each channel of the fourth convolutional feature map to emphasize the regions with higher attention weights. The learned AU feature $\mathbf{f}^{(0)}_{ij}$ with a size of $12c$ is obtained by further applying a convolutional layer and a global average pooling layer~\cite{lin2013network}. To supervise the learning of each global attention map $\mathbf{\widehat{M}}_{ij}$, we use the last one-dimensional fully-connected layer followed by a Sigmoid function to obtain the initially estimated AU occurrence probability $\hat{p}^{(0)}_{ij}$. The AU detection loss\Highlight{~\cite{chen2022geoconv}} for $\hat{p}^{(0)}_{ij}$ with a weighting strategy is defined as
\begin{equation}
\label{eq:L_u0}
\begin{aligned}
&\mathcal{L}_{u}^{(0)}=-\\
&\ \ \frac{1}{t}\sum_{i=1}^t\sum_{j=1}^m w_j [v_j p_{ij} \log \hat{p}^{(0)}_{ij} + (1-p_{ij}) \log (1-\hat{p}^{(0)}_{ij})],
\end{aligned}
\end{equation}
where $w_j$ denotes the weight of the $j$-th AU, $v_j$ denotes the weight for occurrence of the $j$-th AU, and $p_{ij}$ denotes the ground-truth occurrence probability of the $j$-th AU in the $i$-th frame. In the training sets of most AU datasets~\cite{zhang2014bp4d, mavadati2013disfa, girard2017sayette, kollias2019expression,kollias2021analysing}, different AUs have significantly different occurrence rates, and most AUs have much lower occurrence rates than non-occurrence rates. To suppress the two types of data imbalance issues, we define $w_j$ and $v_j$ as
\begin{equation}
    w_j = \frac{n}{n_j^{occ}}/\sum_{k=1}^{m}\frac{n}{n_k^{occ}},\quad v_j=\frac{n-n_j^{occ}}{n_j^{occ}},
\end{equation}
where $n$ and $n_j^{occ}$ are the total number of samples and the number of samples occurring the $j$-th AU in the training set, respectively. The occurrence rate of the $j$-th AU can be formulated as $n_j^{occ}/{n}$.

By combining Eqs.~\eqref{eq:L_attention} and~\eqref{eq:L_u0}, we obtain the full loss of adaptive attention regression network:
\begin{equation}
\label{eq:L_AA}
\mathcal{L}_{AA} = \mathcal{L}_{u}^{(0)} + \lambda_a \mathcal{L}_{a},
\end{equation}
where $\lambda_a$ is used to weigh the importance of the attention regression loss $\mathcal{L}_{a}$. By the constraint of attention predefinition and the guidance of AU detection, the adaptively regressed $\mathbf{\widehat{M}}_{ij}$ can capture both strongly correlated regions \Highlight{with pixel-level dependencies} specified by
landmarks and weakly correlated regions \Highlight{with pixel-level dependencies} distributed globally in the face for subtle and diverse AUs. Due to the regional correlation distribution learned by $\mathbf{\widehat{M}}_{ij}$, the feature $\mathbf{f}^{(0)}_{ij}$ precisely extracts useful information of each AU.

\subsection{Adaptive Spatio-Temporal Graph Convolutional Network}

\highlight{Our goal is to comprehensively reason the correlations in facial AU detection task. Since each AU has specific characteristics, different AUs often have relationships, and adjacent frames often occur or absent similar AUs, we propose an adaptive spatio-temporal graph convolutional network to reason AU-specific patterns, inter-dependencies among AUs, as well as temporal dependencies.} To model the correlations among intra- and inter-frame AUs simultaneously, we combine the features $\mathbf{f}^{(0)}_{ij}$ of $m$ AUs in all the $t$ frames, in which the combined feature $\mathbf{F}^{(0)}$ with size $t\times m\times 12c$ is further fed into the adaptive spatio-temporal graph convolutional network. We first introduce spatial graph convolution without considering temporal operations. 

\subsubsection{Reasoning of AU-Specific Patterns}

Typical graph convolutions are calculated in the spectral domain~\cite{defferrard2016convolutional}, and can be well-approximated by the first-order Chebyshev polynomials expansion~\cite{kipf2017semi}:
\begin{equation}
\label{eq:gcn_ori}
    \mathbf{F}^{out} = (\mathbf{I}+\mathbf{D}^{-\frac{1}{2}}\mathbf{A}\mathbf{D}^{-\frac{1}{2}})\mathbf{F}^{in}\mathbf{\Theta}^{(0)},
\end{equation}
where $\mathbf{F}^{in} \in \mathbb{R}^{m'\times c^{in}}$ and $\mathbf{F}^{out} \in \mathbb{R}^{m'\times c^{out}}$ denote the input and the output, respectively, $\mathbf{I}\in \mathbb{R}^{m'\times m'}$ denotes the identity matrix, $\mathbf{A}\in \mathbb{R}^{m'\times m'}$ denotes the adjacent matrix of the AU relation graph, $\mathbf{D}\in \mathbb{R}^{m'\times m'}$ denotes the degree matrix with $D_{ii} = \sum_j A_{ij}$, and $\mathbf{\Theta}^{(0)}\in \mathbb{R}^{c^{in}\times c^{out}}$ denotes the parameters with omitting the bias. Eq.~\eqref{eq:gcn_ori} essentially transforms $\mathbf{F}^{in}_i\in \mathbb{R}^{1\times c^{in}}$ of the $i$-th AU to $\mathbf{F}^{out}_i\in \mathbb{R}^{1\times c^{out}}$ via the shared $\mathbf{A}$ and $\mathbf{\Theta}^{(0)}$ among all AUs.

Although Eq.~\eqref{eq:gcn_ori} considers the correlations among AUs, it neglects the specific \Highlight{way} of each AU. We propose to share $\mathbf{A}$ to exploit the inter-dependencies among AUs, and apply independent parameters to each AU by using $\mathbf{\Theta}^{(1)}\in \mathbb{R}^{m'\times c^{in}\times c^{out}}$:
\begin{equation}
\label{eq:gcn_new1}
    \mathbf{F}^{out} = (\mathbf{I}+\mathbf{D}^{-\frac{1}{2}}\mathbf{A}\mathbf{D}^{-\frac{1}{2}})\mathbf{F}^{in}\odot\mathbf{\Theta}^{(1)},
\end{equation}  
where the operation ``$\odot$'' in $\mathbf{Z} = \mathbf{X}\odot\mathbf{Y}$ denotes $Z_{ab} = \sum_k X_{ak} Y_{akb}$ at each element $(a,b)$. To reduce the amount of parameters in $\mathbf{\Theta}^{(1)}$, we introduce a disentanglement matrix $\mathbf{Q}\in\mathbb{R}^{m'\times c^e}$ and a parameter pool $\mathbf{W}\in\mathbb{R}^{c^e\times c^{in}\times c^{out}}$ to reformulate the graph convolution as
\begin{equation}
\label{eq:gcn_new2}
    \mathbf{F}^{out} = (\mathbf{I}+\mathbf{D}^{-\frac{1}{2}}\mathbf{A}\mathbf{D}^{-\frac{1}{2}})\mathbf{F}^{in}\odot\mathbf{Q}\mathbf{W},
\end{equation}
where $\mathbf{\Theta}^{(1)}=\mathbf{Q}\mathbf{W}$, and the intermediate dimension $c^e$ is often smaller than $m'$. From the perspective of the $i$-th AU, parameters $\mathbf{\Theta}^{(1)}_i\in \mathbb{R}^{1\times c^{in}\times c^{out}}$ are extracted from the shared $\mathbf{W}$ by the disentanglement matrix $\mathbf{Q}_i\in\mathbb{R}^{1\times c^e}$. The use of $\mathbf{Q}$ and $\mathbf{W}$ is beneficial for reasoning AU-specific patterns.

\subsubsection{Reasoning of Inter-Dependencies among AUs}

Instead of learning $\mathbf{A}$ and further calculating the normalized adjacency matrix $\mathbf{D}^{-\frac{1}{2}}\mathbf{A}\mathbf{D}^{-\frac{1}{2}}$, we directly learn a matrix $\mathbf{U}\in\mathbb{R}^{m'\times c^e}$ to reduce computations:
\begin{equation}
\label{eq:reform_adj}
    \mathbf{D}^{-\frac{1}{2}}\mathbf{A}\mathbf{D}^{-\frac{1}{2}} = \mathcal{N}(\mathcal{R}(\mathbf{U}\mathbf{U}^{\top})),
\end{equation}
where $\mathcal{R}(\mathbf{X})$ as the rectified linear unit (ReLU)~\cite{nair2010rectified} denotes $X_{ab}=\max(X_{ab},0)$ at each element $(a,b)$, and $\mathcal{N}(\mathbf{X})$ for normalization denotes $X_{ab}=X_{ab}/\sum_{a,b}X_{ab}$ at each element $(a,b)$. Eq.~\eqref{eq:reform_adj} adaptively encodes the correlations among AUs such as co-occurrence and mutual exclusion.
Then we can reformulate the graph convolution as
\begin{equation}
\label{eq:gcn_new3}
    \mathbf{F}^{out} = (\mathbf{I}+\mathcal{N}(\mathcal{R}(\mathbf{U}\mathbf{U}^{\top})))\mathbf{F}^{in}\odot\mathbf{Q}\mathbf{W},
\end{equation}
where $\mathbf{U}$ and $\mathbf{W}$ are shared by all AUs for reasoning the inter-dependencies.

\subsubsection{Reasoning of Temporal Dependencies}

\highlight{Each AU lasts for a duration, in which its intensity is similar between adjacent frames and is dynamic across frames.} Gated recurrent unit (GRU)~\cite{cho2014properties} is a popular method for modeling temporal dynamics. A GRU cell consists of an update gate $\mathbf{z}$ and a reset gate $\mathbf{r}$, in which the gating mechanism at time step $\tau$ is defined as the following equations:
\begin{subequations}
\label{eq:gru}
\begin{equation}
    \mathbf{z}_{\tau} = \sigma(\mathbf{W}_z\mathcal{C}(\mathbf{h}_{\tau-1},\mathbf{x}_{\tau})),
\end{equation}    
\begin{equation}
    \mathbf{r}_{\tau} = \sigma(\mathbf{W}_r\mathcal{C}(\mathbf{h}_{\tau-1},\mathbf{x}_{\tau})),
\end{equation}
\begin{equation}
    \widetilde{\mathbf{h}}_{\tau} =\tanh(\mathbf{W}_{\tilde{h}}\mathcal{C}(\mathbf{r}_{\tau}\star \mathbf{h}_{\tau-1},\mathbf{x}_{\tau})),
\end{equation}
\begin{equation}
    \mathbf{h}_{\tau}= \mathbf{z}_{\tau}\star \mathbf{h}_{\tau-1}+(1-\mathbf{z}_{\tau})\star\widetilde{\mathbf{h}}_{\tau},
\end{equation}
\end{subequations}
where $\mathbf{z}_{\tau}$ determines the amount of preserving the previous hidden state $\mathbf{h}_{\tau-1}$, $\mathbf{r}_{\tau}$ determines the combination of the current input $\mathbf{x}_{\tau}$ and the previous hidden state $\mathbf{h}_{\tau-1}$, ``$\star$'' denotes the element-wise multiplication, $\mathcal{C}(\cdot)$ denotes the concatenation, and $\sigma(\cdot)$ and $\tanh(\cdot)$ denote the Sigmoid function and the hyperbolic tangent function, respectively.

By incorporating the spatial graph convolution defined in Eq.~\eqref{eq:gcn_new3} and the GRU defined in Eq.~\eqref{eq:gru}, we formulate our proposed spatio-temporal graph convolutional layer $G^{st}(t',m',c')$ as
\begin{subequations}
\label{eq:g_st}
\begin{equation}
    \mathbf{z}_{\tau} = \sigma((\mathbf{I}+\mathcal{N}(\mathcal{R}(\mathbf{U}\mathbf{U}^{\top})))\mathcal{C}(\mathbf{h}_{\tau-1},\mathbf{F}^{in}_{\tau})\odot\mathbf{Q}\mathbf{W}_z),
\end{equation}    
\begin{equation}
    \mathbf{r}_{\tau} = \sigma((\mathbf{I}+\mathcal{N}(\mathcal{R}(\mathbf{U}\mathbf{U}^{\top})))\mathcal{C}(\mathbf{h}_{\tau-1},\mathbf{F}^{in}_{\tau})\odot\mathbf{Q}\mathbf{W}_r),
\end{equation}
\begin{equation}
\begin{aligned}
    \widetilde{\mathbf{h}}_{\tau} &=\tanh(\\
    &\ (\mathbf{I}+\mathcal{N}(\mathcal{R}(\mathbf{U}\mathbf{U}^{\top})))\mathcal{C}(\mathbf{r}_{\tau}\star \mathbf{h}_{\tau-1},\mathbf{F}^{in}_{\tau})\odot\mathbf{Q}\mathbf{W}_{\tilde{h}}),
\end{aligned}
\end{equation}
\begin{equation}
    \mathbf{h}_{\tau}= \mathbf{z}_{\tau}\star \mathbf{h}_{\tau-1}+(1-\mathbf{z}_{\tau})\star\widetilde{\mathbf{h}}_{\tau},
\end{equation}
\end{subequations}
where $\mathbf{F}^{in}_{\tau}\in \mathbb{R}^{m'\times c'}$ and $\mathbf{h}_{\tau}\in \mathbb{R}^{m'\times c'}$ are the input and the output at time step $\tau$, respectively, 
and $\mathbf{U}\in \mathbb{R}^{m'\times c^{e}}$, $\mathbf{Q}\in \mathbb{R}^{m'\times c^{e}}$, $\mathbf{W}_{z}\in\mathbb{R}^{c^e\times 2c'\times c'}$, $\mathbf{W}_{r}\in\mathbb{R}^{c^e\times 2c'\times c'}$, and $\mathbf{W}_{\tilde{h}}\in\mathbb{R}^{c^e\times 2c'\times c'}$ are learnable parameters. Its elaborated structure is illustrated in Fig.~\ref{fig:sub_blocks}(b), in which $\mathbf{F}^{in}\in \mathbb{R}^{t'\times m'\times c'}$, and $\mathbf{F}^{out}\in \mathbb{R}^{t'\times m'\times c'}$ is the concatenation of $\mathbf{h}_{1},\cdots,\mathbf{h}_{t'}$.

As shown in Fig.~\ref{fig:AAR_framework}, there are two $G^{st}(t,m,12c)$ layers in the adaptive spatio-temporal graph convolutional network to model spatio-temporal correlations among the $m$ AUs in the $t$ frames, in which $\mathbf{U}$, $\mathbf{Q}$, $\mathbf{W}_{z}$, $\mathbf{W}_{r}$, and $\mathbf{W}_{\tilde{h}}$ are not shared in different $G^{st}$ layers. We partition the output feature $\mathbf{F}^{(1)}$ of the last $G^{st}$ layer into a new feature $\mathbf{f}_{ij}^{(1)}$ with a size of $12c$ for each AU in each frame, which is beneficial for final AU detection due to the correlated information from other AUs and frames. Similar to the adaptive attention regression network, we adopt a one-dimensional fully-connected layer followed by a Sigmoid function for each AU to obtain the finally estimated AU occurrence probability $\hat{p}_{ij}^{(1)}$. The AU detection loss for $\hat{p}_{ij}^{(1)}$ is applied to enable the training of our adaptive spatio-temporal graph convolutional network:
\begin{equation}
\label{eq:L_u1}
\begin{aligned}
&\mathcal{L}_{u}^{(1)}=-\\
&\ \frac{1}{t}\sum_{i=1}^t\sum_{j=1}^m w_j [v_j p_{ij} \log \hat{p}^{(1)}_{ij} + (1-p_{ij}) \log (1-\hat{p}^{(1)}_{ij})].
\end{aligned}
\end{equation}
At test time, our framework can predict the AU occurrence probabilities of an input sequence with any number of frames, which has flexible applicability.

\begin{table*}
\centering\caption{AU occurrence rates ($\%$) in the training sets of different datasets. ``-'' denotes the AU is not annotated in this dataset.}
\label{tab:au_occurrence}
\begin{tabular}{c|*{15}{c}}
\toprule
AU &1 &2 &4 &6 &7 &9 &10 &12 &14 &15 &17 &23 &24 &25 &26\\
\midrule
BP4D~\cite{zhang2014bp4d} &21.1 &17.1 &20.3 &46.2 &54.9 &- &59.4 &56.2 &46.6 &16.9 &34.4 &16.5 &15.2 &- &-\\
DISFA~\cite{mavadati2013disfa} &5.0 &4.0 &15.0 &8.1 &- &4.3 &- &13.2 &- &- &- &- &- &27.8 &8.9\\
GFT~\cite{girard2017sayette} &3.7 &13.5 &3.7 &28.3 &- &- &24.6 &29.3 &3.1 &10.7 &- &25.0 &14.1 &- &- \\
Aff-Wild2~\cite{kollias2019expression,kollias2021analysing} &11.9 &5.1 &16.0 &26.5 &39.9 &- &34.5 &24.3 &- &2.8 &- &3.1 &2.8 &62.8 &7.6\\
\bottomrule
\end{tabular}
\end{table*}

\begin{table*}
\centering\caption{F1-frame results for $12$ AUs on BP4D~\cite{zhang2014bp4d}. The reported results of previous methods are from their original papers.}
\label{tab:comp_f1_bp4d}
\setlength\tabcolsep{8pt}
\begin{tabular}{c|*{13}{c}}
\toprule
AU &1 &2 &4 &6 &7 &10 &12 &14 &15 &17 &23 &24 &\textbf{Avg}\\
\midrule
DRML~\cite{zhao2016deep} &36.4 &41.8 &43.0 &55.0 &67.0 &66.3 &65.8 &54.1 &33.2 &48.0 &31.7 &30.0 &48.3\\
EAC-Net~\cite{li2018eac} &39.0 &35.2 &48.6 &76.1 &72.9 &81.9 &86.2 &58.8 &37.5 &59.1 &35.9 &35.8 &55.9\\
DSIN~\cite{corneanu2018deep}  &51.7 &40.4 &56.0 &76.1 &73.5 &79.9 &85.4 &62.7 &37.3 &62.9 &38.8 &41.6 &58.9\\
CMS~\cite{sankaran2019representation} &49.1 &44.1 &50.3 &79.2 &74.7 &80.9 &88.3 &63.9 &44.4 &60.3 &41.4 &51.2 &60.6\\
LP-Net~\cite{niu2019local} &43.4 &38.0 &54.2 &77.1 &76.7 &83.8 &87.2 &63.3 &45.3 &60.5 &48.1 &54.2 &61.0\\
SRERL~\cite{li2019semantic} &46.9 &45.3 &55.6 &77.1 &78.4 &83.5 &87.6 &60.6 &52.2 &63.9 &47.1 &53.3 &62.9\\
ARL~\cite{shao2019facial} &45.8 &39.8 &55.1 &75.7 &77.2 &82.3 &86.6 &58.8 &47.6 &62.1 &47.4 &55.4 &61.1\\
AU R-CNN~\cite{ma2019r} &50.2 &43.7 &57.0 &78.5 &78.5 &82.6 &87.0 &\textbf{67.7} &49.1 &62.4 &50.4 &49.3 &63.0\\
AU-GCN~\cite{liu2020relation} &46.8 &38.5 &60.1 &\textbf{80.1} &79.5 &84.8 &88.0 &67.3 &52.0 &63.2 &40.9 &52.8 &62.8\\
JÂA-Net~\cite{shao2021jaa}  &53.8 &47.8 &58.2 &78.5 &75.8 &82.7 &88.2 &63.7 &43.3 &61.8 &45.6 &49.9 &62.4\\
UGN-B~\cite{song2021uncertain} &54.2 &46.4 &56.8 &76.2 &76.7 &82.4 &86.1 &64.7 &51.2 &63.1 &48.5 &53.6 &63.3\\
HMP-PS~\cite{song2021hybrid} &53.1 &46.1 &56.0 &76.5 &76.9 &82.1 &86.4 &64.8 &51.5 &63.0 &49.9 &54.5 &63.4\\
\highlight{Jacob \etal~\cite{jacob2021facial}} &51.7 &49.3 &\textbf{61.0} &77.8 &79.5 &82.9 &86.3 &67.6 &51.9 &63.0 &43.7 &56.3 &64.2\\
\highlight{RTATL~\cite{yan2021self}} &\textbf{57.1} &\textbf{49.7} &60.5 &77.9 &76.1 &84.4 &87.2 &64.3 &53.5 &\textbf{67.0} &48.9 &48.6 &64.6\\
\highlight{Li \etal~\cite{li2021integrating}} &54.0 &46.0 &55.7 &79.4 &78.8 &84.5 &87.0 &67.0 &55.6 &63.1 &50.7 &55.3 &\textbf{64.8}\\
\highlight{Chang \etal~\cite{chang2022knowledge}} &53.3 &47.4 &56.2 &79.4 &\textbf{80.7} &\textbf{85.1} &\textbf{89.0} &67.4 &\textbf{55.9} &61.9 &48.5 &49.0 &64.5\\
\textbf{AAR} &53.2 &47.7 &56.7 &75.9 &79.1 &82.9 &88.6 &60.5 &51.5 &61.9 &\textbf{51.0} &\textbf{56.8} &63.8\\
\bottomrule
\end{tabular}
\end{table*}

\section{Experiments}

\subsection{Datasets and Settings}

\subsubsection{Datasets}

We evaluate our AAR on four benchmark datasets, including BP4D~\cite{zhang2014bp4d}, DISFA~\cite{mavadati2013disfa}, and GFT~\cite{girard2017sayette} in constrained scenarios, and Aff-Wild2~\cite{kollias2019expression,kollias2021analysing} in unconstrained scenarios. The AU labels of each dataset is annotated by certified FACS experts. 
\begin{itemize}
    \item\textbf{BP4D} consists of $41$ subjects with $23$ females and $18$ males, in which each subject participates in $8$ sessions. There are totally $328$ videos including about $140,000$ frames, each of which is annotated with AUs of occurrence or non-occurrence, as well as $49$ facial landmarks detected by SDM~\cite{xiong2013supervised}. Similar to previous works~\cite{zhao2016deep,li2018eac,shao2021jaa}, we conduct subject-exclusive 3-fold cross-validation on $12$ AUs: 1, 2, 4, 6, 7, 10, 12, 14, 15, 17, 23 and 24, in which two folds are used for training and the rest one is used for testing.
    \item\textbf{DISFA} includes $12$ female and $15$ male subjects, each of whom is recorded by a video with $4,845$ frames. Each frame is annotated with AU intensities ranging from $0$ to $5$, as well as $66$ landmarks detected by AAM~\cite{cootes2001active}. Following the settings of previous works~\cite{zhao2016deep,li2018eac,shao2021jaa}, we treat an AU as occurrence if its intensity is equal or greater than $2$ and treat it as non-occurrence otherwise. We also conduct a subject-exclusive 3-fold cross-validation on $8$ AUs: 1, 2, 4, 6, 9, 12, 25 and 26.
    \item\textbf{GFT} contains $96$ subjects in $32$ three-subject groups with unscripted conversations, in which each subject is recorded by a video with most frames in moderate out-of-plane poses. Each frame is annotated with $10$ AUs (1, 2, 4, 6, 10, 12, 14, 15, 23, and 24), as well as $49$ landmarks detected by ZFace~\cite{jeni2017dense}. Following the official training and testing partitions~\cite{girard2017sayette}, we use $78$ subjects with about $108,000$ frames for training, and $18$ subjects with about $24,600$ frames for testing.
    \item\textbf{Aff-Wild2} is a large-scale in-the-wild facial expression dataset, whose videos are collected from YouTube with a wide variability in ages, ethnicities, professions, emotions, poses, illumination, and occlusions. 
    It consists of $305$ videos with about $1,390,000$ frames as a training set, and $105$ videos with about $440,000$ frames as a validation set. Each frame is annotated with $12$ AUs: 1, 2, 4, 6, 7, 10, 12, 15, 23, 24, 25, and 26. Since facial landmarks are not annotated, we adopt a powerful landmark detection library Dlib~\cite{dlib09,kazemi2014one} to detect $68$ landmarks of each frame, \Highlight{in which the landmark annotations will be released}. Similar to Zhang \etal~\cite{zhang2021prior}, we train our AAR on the training set, and test on the validation set.
\end{itemize}

The AU occurrence rates in the training sets of these datasets are shown in Table~\ref{tab:au_occurrence}, in which data imbalance issues exist in each dataset.

\subsubsection{Implementation Details} 

Our AAR is implemented based on PyTorch~\cite{paszke2019pytorch}, in which all the convolutional layers employ $3\times 3$ kernels with stride $1$ and padding $1$, and all the max-pooling layers process $2\times 2$ spatial fields with stride $2$. As with the previous techniques~\cite{shao2021jaa,shao2019facial}, each image is aligned to $3\times 200\times 200$ using similarity transformation \highlight{by fitting facial landmarks}, and is further randomly cropped to $3\times l\times l$ and mirrored. \highlight{To preserve the consistency across frames, all the $t$ frames in a specific input sequence are processed with identical cropping and identical mirroring.} The crop size $l$, the standard deviation $\delta$, the trade-off parameter $\lambda_a$, the number of frames $t$ in an input training sequence, and the intermediate dimension $c^e$ are set to $176$, $3$, $2$, $48$, and $8$, respectively. The number of AUs $m$ is $12$, $8$, $10$, and $12$ for BP4D, DISFA, GFT, and Aff-Wild2, respectively.

We employ the stochastic gradient descent (SGD) solver, a Nesterov momentum~\cite{sutskever2013importance} of $0.9$, and a weight decay of $0.0005$ to train the adaptive attention regression network for up to $12$ epochs, in which the learning rate with initial value of $0.006$ is multiplied by a factor of $0.3$ at every $2$ epochs. Then we employ the Adam solver~\cite{kingma2014adam} with fixed learning rate $0.006$, and set $\beta_1=0.9$ and $\beta_2=0.999$ to train the adaptive spatio-temporal graph convolutional network for up to $12$ epochs. Following the settings in~\cite{zhao2016deep,li2018eac,shao2021jaa}, AAR model trained on DISFA is initialized using the parameters of the well-trained model on BP4D.


\subsubsection{Evaluation Metrics}

We report the commonly used metric, frame-based F1-score (F1-frame), for AU detection:
$F1=2PR/(P+R)$, where $P$ and $R$ denote the precision and the recall, respectively. The average results of F1-frame over all AUs (abbreviated as Avg) are also shown. In the following sections, all the F1-frame results are reported in percentage with ``$\%$'' omitted.

\begin{table*}
\centering\caption{F1-frame results for $8$ AUs on DISFA~\cite{mavadati2013disfa}.}
\label{tab:comp_f1_disfa}
\setlength\tabcolsep{14pt}
\begin{tabular}{c|*{9}{c}}
\toprule
AU &1 &2 &4 &6 &9 &12 &25 &26 &\textbf{Avg}\\
\midrule
DRML~\cite{zhao2016deep} &17.3 &17.7 &37.4 &29.0 &10.7 &37.7 &38.5 &20.1 &26.7\\
EAC-Net~\cite{li2018eac} &41.5 &26.4 &66.4 &50.7 &8.5 &\textbf{89.3} &88.9 &15.6 &48.5\\
DSIN~\cite{corneanu2018deep}  &42.4 &39.0 &68.4 &28.6 &46.8 &70.8 &90.4 &42.2 &53.6\\
CMS~\cite{sankaran2019representation} &40.2 &44.3 &53.2 &57.1 &50.3 &73.5 &81.1 &59.7 &57.4\\
LP-Net~\cite{niu2019local} &29.9 &24.7 &72.7 &46.8 &49.6 &72.9 &93.8 &65.0 &56.9\\
SRERL~\cite{li2019semantic} &45.7 &47.8 &59.6 &47.1 &45.6 &73.5 &84.3 &43.6 &55.9\\
ARL~\cite{shao2019facial} &43.9 &42.1 &63.6 &41.8 &40.0 &76.2 &\textbf{95.2} &66.8 &58.7\\
AU R-CNN~\cite{ma2019r} &32.1 &25.9 &59.8 &55.3 &39.8 &67.7 &77.4 &52.6 &51.3\\
AU-GCN~\cite{liu2020relation} &32.3 &19.5 &55.7 &\textbf{57.9} &\textbf{61.4} &62.7 &90.9 &60.0 &55.0\\
JÂA-Net~\cite{shao2021jaa}  &\textbf{62.4} &\textbf{60.7} &67.1 &41.1 &45.1 &73.5 &90.9 &67.4 &63.5\\
UGN-B~\cite{song2021uncertain} &43.3 &48.1 &63.4 &49.5 &48.2 &72.9 &90.8 &59.0 &60.0\\
HMP-PS~\cite{song2021hybrid} &38.0 &45.9 &65.2 &50.9 &50.8 &76.0 &93.3 &67.6 &61.0\\
\highlight{Jacob \etal~\cite{jacob2021facial}}&46.1 &48.6 &\textbf{72.8} &56.7 &50.0 &72.1 &90.8 &55.4 &61.5\\
\highlight{RTATL~\cite{yan2021self}} &57.8 &52.8 &70.8 &53.2 &52.7 &74.5 &91.5 &51.9 &63.1\\
\highlight{Li \etal~\cite{li2021integrating}} &47.5 &53.3 &64.4 &51.8 &44.4 &74.7 &92.1 &60.7 &61.1\\
\highlight{Chang \etal~\cite{chang2022knowledge}} &60.4 &59.2 &67.5 &52.7 &51.5 &76.1 &91.3 &57.7 &\textbf{64.5}\\
\textbf{AAR} &\textbf{62.4} &53.6 &71.5 &39.0 &48.8 &76.1 &91.3 &\textbf{70.6} &64.2\\
\bottomrule
\end{tabular}
\end{table*}

\begin{table*}
\centering\caption{F1-frame results for $10$ AUs on GFT~\cite{girard2017sayette}. The reported results of LSVM~\cite{fan2008liblinear} and AlexNet~\cite{krizhevsky2012imagenet} are from~\cite{girard2017sayette}, and those of EAC-Net~\cite{li2018eac} and ARL~\cite{shao2019facial} are from~\cite{shao2021jaa}.}
\label{tab:comp_f1_gft}
\setlength\tabcolsep{10.4pt}
\begin{tabular}{c|*{10}{c}|c}
\toprule
AU&1&2&4&6&10&12&14&15&23&24&Avg\\
\midrule
LSVM~\cite{fan2008liblinear} &38 &32 &13 &67 &64
&78 &15 &29 &49 &44 &42.9\\
AlexNet~\cite{krizhevsky2012imagenet} &44 &46 &2 &73 &72 &82 &5 &19 &43 &42 &42.8\\
EAC-Net~\cite{li2018eac} &15.5 &\textbf{56.6} &0.1 &81.0 &76.1 &84.0 &0.1 &38.5 &57.8 &\textbf{51.2} &46.1\\
TCAE~\cite{li2019self-supervised} &43.9 &49.5 &6.3
&71.0 &76.2 &79.5 &10.7 &28.5 &34.5 &41.7 &44.2\\
ARL~\cite{shao2019facial} &51.9 &45.9 &13.7 &79.2 &75.5 &82.8 &0.1 &44.9 &\textbf{59.2} &47.5 &50.1\\
Ertugrul \etal~\cite{ertugrul2020crossing} &43.7 &44.9 &19.8 &74.6 &\textbf{76.5} &79.8 &\textbf{50.0} &33.9 &16.8 &12.9 &45.3\\
JÂA-Net~\cite{shao2021jaa} &46.5 &49.3 &19.2 &79.0 &75.0 &\textbf{84.8} &44.1 &33.5 &54.9 &50.7 &53.7\\
\textbf{AAR} &\textbf{66.3} &53.9 &\textbf{23.7} &\textbf{81.5} &73.6 &84.2 &43.8 &\textbf{53.8} &58.2 &46.5 &\textbf{58.5}\\
\bottomrule
\end{tabular}
\end{table*}

\begin{table*}
\centering\caption{F1-frame results for $12$ AUs on Aff-Wild2~\cite{kollias2019expression,kollias2021analysing}. EAC-Net~\cite{li2018eac} is implemented on Aff-Wild2 using its released code.}
\label{tab:comp_f1_affwild2}
\setlength\tabcolsep{8.3pt}
\begin{tabular}{c|*{12}{c}|c}
\toprule
AU &1 &2 &4 &6 &7 &10 &12 &15 &23 &24 &25 &26 &Avg\\
\midrule
EAC-Net~\cite{li2018eac} &49.6 &33.7 &55.6 &66.4 &82.3 &81.4 &76.9 &11.8 &12.5 &12.2 &93.7 &26.8 &50.2\\
ARL~\cite{shao2019facial} &59.2 &48.2 &54.9 &70.0 &83.4 &80.3 &72.0 &0.1 &0.1 &17.3 &93.0 &37.5 &51.3\\
JÂA-Net~\cite{shao2021jaa} &61.7 &50.1 &56.0 &71.7 &81.7 &82.3 &78.0 &\textbf{31.1} &1.4 &8.6 &\textbf{94.8} &37.5 &54.6\\
Zhang \etal~\cite{zhang2021prior} &\textbf{65.7} &\textbf{64.2} &\textbf{66.5} &\textbf{76.6} &74.7 &72.7 &78.6 &18.5 &10.6 &\textbf{55.1} &80.7 &\textbf{41.7} &58.8\\
\textbf{AAR} &65.4 &57.9 &59.9 &73.2 &\textbf{84.6} &\textbf{83.2} &\textbf{79.9} &21.8 &\textbf{27.4} &19.9 &94.5 &\textbf{41.7} &\textbf{59.1}\\
\bottomrule
\end{tabular}
\end{table*}

\subsection{Comparison with State-of-the-Art Methods}

We compare our AAR against state-of-the-art AU detection methods under the same evaluation setting. These methods include LSVM~\cite{fan2008liblinear}, AlexNet~\cite{krizhevsky2012imagenet}, DRML~\cite{zhao2016deep}, EAC-Net~\cite{li2018eac}, DSIN~\cite{corneanu2018deep}, CMS~\cite{sankaran2019representation}, LP-Net~\cite{niu2019local}, SRERL~\cite{li2019semantic}, ARL~\cite{shao2019facial}, AU R-CNN~\cite{ma2019r}, TCAE~\cite{li2019self-supervised}, AU-GCN~\cite{liu2020relation}, Ertugrul \etal~\cite{ertugrul2020crossing}, JÂA-Net~\cite{shao2021jaa}, UGN-B~\cite{song2021uncertain}, HMP-PS~\cite{song2021hybrid}, Zhang \etal~\cite{zhang2021prior}, \highlight{Jacob \etal~\cite{jacob2021facial}, RTATL~\cite{yan2021self}, Li \etal~\cite{li2021integrating}, and Chang \etal~\cite{chang2022knowledge}. Our AAR, RTATL, and Li \etal~\cite{li2021integrating} employ temporal information, and all the remaining methods process a single frame at a time without considering temporal information.} Note that most of these approaches use external training data, while our method only uses the benchmark dataset. Specifically, EAC-Net, SRERL, AU R-CNN, UGN-B, HMP-PS, \highlight{Jacob \etal~\cite{jacob2021facial}, RTATL, Li \etal~\cite{li2021integrating}, and Chang \etal~\cite{chang2022knowledge}} fine-tune the official pre-trained VGG~\cite{simonyan2014very}, ResNet~\cite{he2016deep}, or \highlight{InceptionV3~\cite{szegedy2016rethinking}} models, CMS utilizes external thermal images to guide the representation learning, LP-Net pre-trains its network on a face recognition dataset VGGFace2~\cite{cao2018vggface2}, and Zhang \etal~\cite{zhang2021prior} employs BP4D~\cite{zhang2014bp4d} when trained on Aff-Wild2~\cite{kollias2019expression,kollias2021analysing}.

\subsubsection{Evaluation on BP4D}
Table~\ref{tab:comp_f1_bp4d} shows the F1-frame results of our method AAR and state-of-the-art methods on the BP4D benchmark. We can see that our AAR with average F1-frame $63.8$ outperforms \highlight{most of} the previous methods. Compared to the methods like UGN-B, \highlight{Jacob \etal~\cite{jacob2021facial}, and Chang \etal~\cite{chang2022knowledge}} using additional training images, AAR achieves \highlight{comparable} performance with only benchmark training images, which demonstrates our framework has a strong capability of suppressing the model overfitting. Note that AAR performs better than SRERL, AU-GCN, UGN-B, and HMP-PS which also exploit graph neural networks. The gain comes from the accurate regional localization by our proposed adaptive attention regression network and the complete relational reasoning by our proposed adaptive spatio-temporal graph convolutional network.

\subsubsection{Evaluation on DISFA}

We report the F1-frame results on DISFA in Table~\ref{tab:comp_f1_disfa}. It can be observed that our AAR outperforms most approaches by large margins. 
Moreover, there is a more serious data imbalance issue in DISFA than BP4D, as shown in Table~\ref{tab:au_occurrence}, which causes large performance fluctuations over different AUs for most of the previous approaches such as EAC-Net and AU-GCN. In this case, our AAR achieves more stable performance across AUs. By comparing Table~\ref{tab:comp_f1_bp4d} and Table~\ref{tab:comp_f1_disfa}, we can see that a few recent works like AU R-CNN and \highlight{Li \etal~\cite{li2021integrating} work well on BP4D while achieve mediocre performance on DISFA}, which suggests their limited applicability. In contrast, AAR obtains consistently good performance on BP4D and DISFA. \highlight{Besides, compared to the most recent video based works RTATL and Li \etal~\cite{li2021integrating}, AAR works better on DISFA and performs worse on BP4D. This is partly due to the use of external training data in RTATL and Li \etal~\cite{li2021integrating}.}

\subsubsection{Evaluation on GFT}

Table~\ref{tab:comp_f1_gft} presents the F1-frame results on GFT, in which our AAR exhibits significantly better performance than previous works. Compared with BP4D and DISFA whose images are near-frontal faces, GFT contains many images under out-of-plane poses. In this challenging case, other methods like EAC-Net fail to work well. In contrast, AAR achieves good performance with average F1-frame $58.5$.

\begin{table*}
\centering\caption{The structures of different variants of our AAR. \textbf{HM}: two hierarchical and multi-scale region layers. \textbf{C}: five stacked convolutional layers in each AU branch. $W^{au}$: weighting strategy in AU detection loss. \textbf{AL}: attention learning in each AU branch. $G^{sh}$: spatial graph convolution using Eq.~\eqref{eq:gcn_new_tmp}. $G^{sp}$: spatial graph convolution using Eq.~\eqref{eq:gcn_new3}. $T^{avg}$: treating the average predictions over all $t$ frames in the input sequence as the prediction of each frame. $T^{gru}$: reasoning temporal dependencies using Eq.~\eqref{eq:gru}.}
\label{tab:variant_AAR}
\setlength\tabcolsep{10.7pt}
\begin{tabular}{c|*{11}{c}}
\toprule
Method &HM &C &$W^{au}$ &AL &$G^{sh}$ &$G^{sp}$ &$T^{avg}$ &$T^{gru}$ &$\mathcal{L}_{a}$ &$\mathcal{L}_{u}^{(0)}$ &$\mathcal{L}_{u}^{(1)}$\\
\midrule
B-Net &$\surd$ &$\surd$ & & & & & & & &$\surd$ &\\
BW-Net &$\surd$ &$\surd$ &$\surd$ & & & & & & &$\surd$ &\\
AA &$\surd$ &$\surd$ &$\surd$ &$\surd$ & & & & &$\surd$ &$\surd$ &\\
AA+$G^{sh}$ &$\surd$ &$\surd$ &$\surd$ &$\surd$ &$\surd$ & & & &$\surd$ &$\surd$ &$\surd$\\
AA+$G^{sp}$ &$\surd$ &$\surd$ &$\surd$ &$\surd$ & &$\surd$ & & &$\surd$ &$\surd$ &$\surd$\\
AA+$G^{sp}$+$T^{avg}$ &$\surd$ &$\surd$ &$\surd$ &$\surd$ & &$\surd$ &$\surd$ & &$\surd$ &$\surd$ &$\surd$\\
\textbf{AAR}&$\surd$ &$\surd$ &$\surd$ &$\surd$ & &$\surd$ & &$\surd$ &$\surd$ &$\surd$ &$\surd$\\
\bottomrule
\end{tabular}
\end{table*}

\begin{table*}
\centering\caption{F1-frame results for $12$ AUs of different variants of AAR on BP4D~\cite{zhang2014bp4d}.}
\label{tab:ablation_bp4d}
\setlength\tabcolsep{8.3pt}
\begin{tabular}{c|*{12}{c}|c}
\toprule
AU &1 &2 &4 &6 &7 &10 &12 &14 &15 &17 &23 &24 &Avg\\
\midrule
B-Net &47.8 &42.1 &51.4 &72.6 &73.4 &79.5 &85.6 &58.3 &45.3 &59.9 &40.8 &48.8 &58.8\\

BW-Net &49.0 &44.2 &52.9 &73.9 &74.9 &79.3 &84.5 &59.1 &48.0 &61.1 &41.8 &50.4 &59.9\\

AA &48.8 &45.7 &53.9 &73.0 &77.6 &82.5 &86.8 &59.9 &49.8 &59.1 &48.6 &53.9 &61.6\\
AA+$G^{sh}$ &49.9 &45.3 &54.9 &74.5 &77.5 &81.7 &87.5 &59.3 &49.6 &59.5 &47.5 &53.7 &61.7\\
AA+$G^{sp}$ &50.9 &46.2 &55.0 &74.9 &\textbf{79.1} &82.7 &87.8 &\textbf{62.7} &51.1 &59.5 &49.3 &54.7 &62.8\\
AA+$G^{sp}$+$T^{avg}$ &49.8 &45.7 &53.1 &74.3 &78.8 &82.6 &88.4 &60.8 &45.8 &54.8 &38.8 &49.7 &60.2\\
\textbf{AAR} &\textbf{53.2} &\textbf{47.7} &\textbf{56.7} &\textbf{75.9} &\textbf{79.1} &\textbf{82.9} &\textbf{88.6} &60.5 &\textbf{51.5} &\textbf{61.9} &\textbf{51.0} &\textbf{56.8} &\textbf{63.8}\\
\bottomrule
\end{tabular}
\end{table*}

\subsubsection{Evaluation on Aff-Wild2}

We have validated the performance of our AAR in constrained scenarios. To also investigate the effectiveness in unconstrained scenarios, we compare with other works on the challenging Aff-Wild2, as shown in Table~\ref{tab:comp_f1_affwild2}. It can be seen that our AAR has higher average F1-frame than previous methods. Note that EAC-Net and Zhang \etal~\cite{zhang2021prior} exploit external training data, while AAR still achieves the overall best performance by only using Aff-Wild2 dataset. \highlight{By observing the results of JÂA-Net and AAR from Table~\ref{tab:comp_f1_bp4d} to Table~\ref{tab:comp_f1_affwild2}, we can find that the margins between JÂA-Net and AAR on GFT and Aff-Wild2 are larger than those on BP4D and DISFA.
This demonstrates that AAR can more effectively process challenging cases in the detection of AUs.}

\subsection{Ablation Study}

In this section, we evaluate each component in our AAR framework. The structures of different variants of AAR are summarized in Table~\ref{tab:variant_AAR}. B-Net is a baseline method, which only contains the adaptive attention regression network without the structure of learning the attention map $\mathbf{\widehat{M}}_{ij}$. Besides, it does not employ the weighting strategy in Eq.~\eqref{eq:L_u0} by setting $w_j=1/m$ and $v_j=1$. AA is the adaptive attention regression network with the full loss defined in Eq.~\eqref{eq:L_AA}. $G^{sh}$ denotes sharing $\mathbf{\Theta}^{(0)}$ among all AUs in the spatial graph convolution:
\begin{equation}
\label{eq:gcn_new_tmp}
    \mathbf{F}^{out} = (\mathbf{I}+\mathcal{N}(\mathcal{R}(\mathbf{U}\mathbf{U}^{\top})))\mathbf{F}^{in} \mathbf{\Theta}^{(0)},
\end{equation}
while $G^{sp}$ reasons AU-specific patterns by introducing $\mathbf{Q}$ and $\mathbf{W}$ in Eq.~\eqref{eq:gcn_new3}. $T^{avg}$ is a plain temporal method, which uses the average predictions over all $t$ frames in the input sequence as the prediction of each frame. Table~\ref{tab:ablation_bp4d} presents the F1-frame results of different variants of AAR on BP4D. 

\subsubsection{Weighting Strategy for Suppressing Data Imbalance}

After using $w_j$ to suppress the data imbalance that different AUs have significantly different occurrence rates, and using $v_j$ to suppress the data imbalance that most AUs have much lower occurrence rates than non-occurrence rates, BW-Net achieves higher average F1-frame $59.9$ than B-Net. Since inter-AU and intra-AU data imbalance issues are both suppressed, the weighting strategy can contribute to better AU detection performance. 

\subsubsection{Adaptive Attention Regression}

Compared with BW-Net, AA adaptively regresses the attention map of each AU under the constraint of attention predefinition and the guidance of AU detection. We can see that AA significantly improves the average F1-frame from $59.9$ to $61.6$. This demonstrates the effectiveness of our proposed adaptive attention regression network.

\subsubsection{Reasoning of AU-Specific Patterns}

In Table~\ref{tab:ablation_bp4d}, we can observe AA+$G^{sh}$ improves the results slightly over AA. This is because the patterns of different AUs are often different, which should not be reasoned with the shared parameters $\mathbf{\Theta}^{(0)}$. In contrast, by using Eq.~\eqref{eq:gcn_new3} to reason AU-specific patterns, AA+$G^{sp}$ improves the average F1-frame from $61.6$ to $62.8$ over AA.

\subsubsection{Reasoning of Temporal Dependencies}

We implement AA+$G^{sp}$+$T^{avg}$ by calculating the average predictions of AA+$G^{sp}$ over frames as the final prediction of each frame. It can be seen that the results become significantly worse. This indicates the correlations among frames are non-trivial, which should be captured by a powerful temporal model. After reasoning the dependencies among frames by GRU, our AAR achieves the best results on the BP4D benchmark. The independent pattern of each AU, the inter-dependencies among AUs, as well as the temporal dependencies are simultaneously modeled by our adaptive spatio-temporal graph convolutional network, in which spatial and temporal relevant information are both exploited to improve the performance of AU detection. In our framework, the adaptive attention regression network contributes to extracting accurate AU features, and the adaptive spatio-temporal graph convolutional network reasons complete correlations based on the AU features to achieve the final performance.

\begin{figure*}
\centering\includegraphics[width=\linewidth]{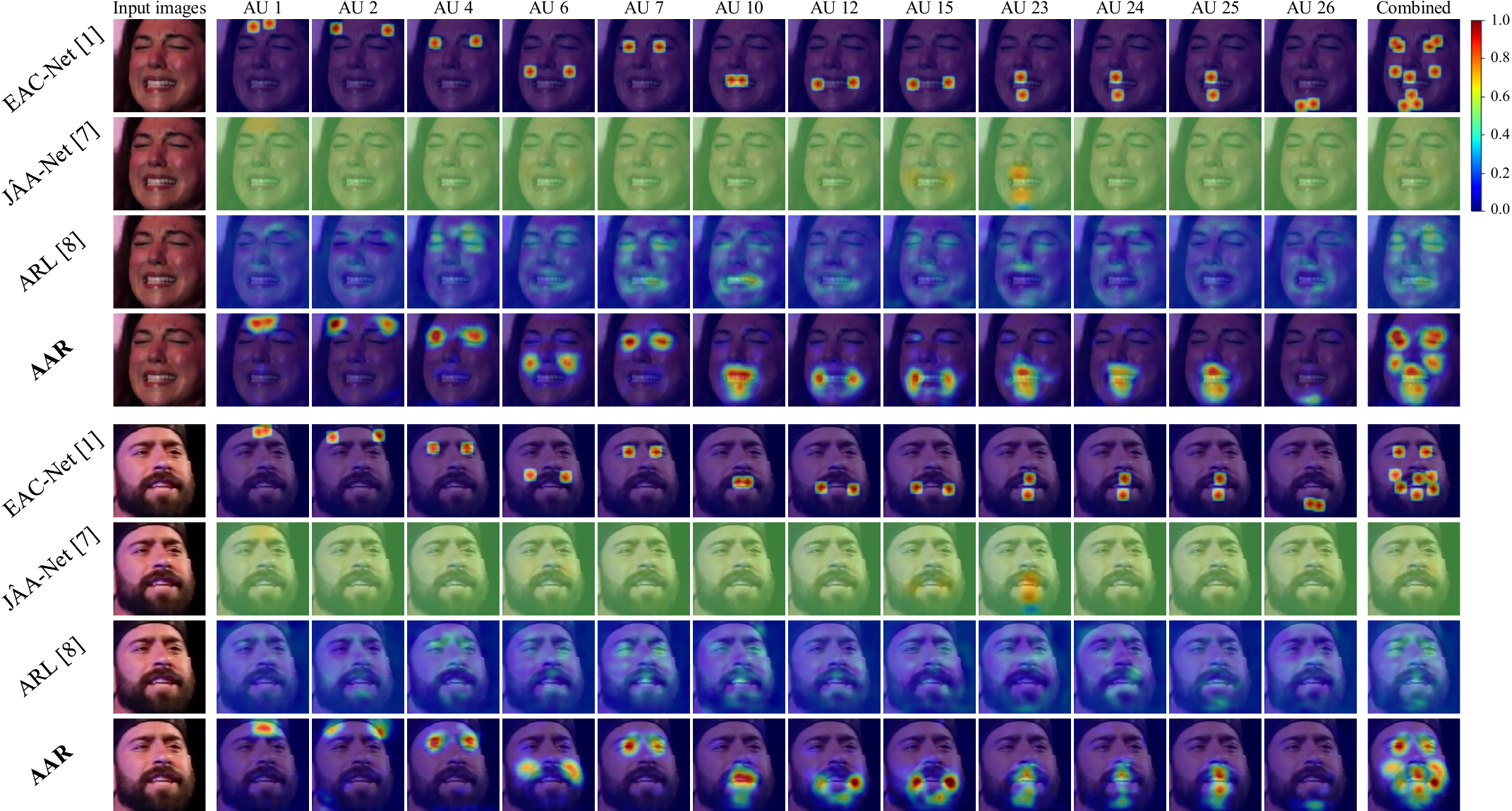}
\caption{Visualization of the learned attention maps by different methods for two example Aff-Wild2~\cite{kollias2019expression,kollias2021analysing} images, in which the first image occurs AUs 4, 6, 7, 25, and 26, and the second image occurs AUs 6, 7, 12, and 25. Each row shows the attention maps of $12$ AUs as well as the combined attention map of occurred AUs for one method. Attention weights from $0$ to $1$ are visualized with different colors in the color bar, which are overlaid on the input images for better viewing.}
\label{fig:attention_map}
\end{figure*}

\subsection{Visual Results}

Fig.~\ref{fig:attention_map} shows the comparisons of the learned attention maps for our AAR and recent attention learning based methods EAC-Net~\cite{li2018eac}, JÂA-Net~\cite{shao2021jaa}, and ARL~\cite{shao2019facial}. We can observe that the attentions of each AU in EAC-Net have a fixed distribution and only exist in the predefined ROIs with a fixed size, which ignores the potentially correlated regions \Highlight{with pixel-level dependencies} beyond the ROIs. Although JÂA-Net tries to refine the attentions predefined in EAC-Net, it seems that the refined attentions are the smoothing of the predefined attentions, in which the regions far away from the ROIs obtain equal attention weights. Since relevant and irrelevant regions included in these regions are treated with equal importance, the extraction of AU features may be not accurate. In this case, the learned attentions of the same AUs in different images look similar. However, the regional correlation distribution of each AU should be diverse across persons and expressions. For example, AUs 6, 7, and 25 co-occur with AUs 4 and 26 in the first image, while co-occur with AU 12 in the second image.

On the other hand, we can see that the learned attentions of ARL almost contain all correlated regions. However, quite a few irrelevant regions are also included, and the strongly correlated regions \Highlight{with pixel-level dependencies} around the AU centers are sometimes not fully captured. In contrast, our AAR can capture both strongly correlated regions \Highlight{with pixel-level dependencies} predefined by landmarks and weakly correlated regions \Highlight{with pixel-level dependencies} distributed globally in the face. \highlight{For example, AU 1 (inner brow raiser) and AU 10 (upper lip raiser) often co-occur, in which the learned attention map of AU 1 by AAR has large attention weights in the inner brow regions and has small attention weights around the upper lip regions, and the learned attention map of AU 10 by AAR has large attention weights in the lip regions and has small attention weights around the brow regions. AU 6 (cheek raiser) and AU 7 (lid tightener) also often co-occur, and we can easily observe a similar phenomenon.} We also notice that the captured correlated regions of each AU are within the attention range of the combined attention map of occurred AUs. This indicates the irrelevant regions of each AU are accurately discarded. Our method is able to accurately learn the regional correlation distribution of each AU.

\highlight{There are two main reasons for the different observations between our AAR and the two previous adaptive attention based methods JÂA-Net~\cite{shao2021jaa} and ARL~\cite{shao2019facial}:
\begin{itemize}
    \item The adaptive attention learning of JÂA-Net is achieved by stacked convolutions on a predefined attention map of each AU. Since the predefined attention map has a fixed attention distribution across samples, directly convoluting on a single-channel map with a fixed value distribution has a limited refinement space. In contrast, the attention map of each AU in AAR is learned from features, which are adaptively learned and are diverse for different samples. AAR can adaptively learn attention maps according to the characteristics of each sample.
    \item The main difference of adaptive attention learning between ARL and AAR is the discarding of prior knowledge about AU locations in ARL. Compared to the use of the fully-connected conditional random field (CRF)~\cite{krahenbuhl2011efficient,zheng2015conditional} in ARL to capture pixel-level correlations, the constraint of attention predefinition in AAR can more accurately remove the highlighting of irrelevant regions in attention maps.
\end{itemize}
}

\begin{figure*}
\centering\includegraphics[width=\linewidth]{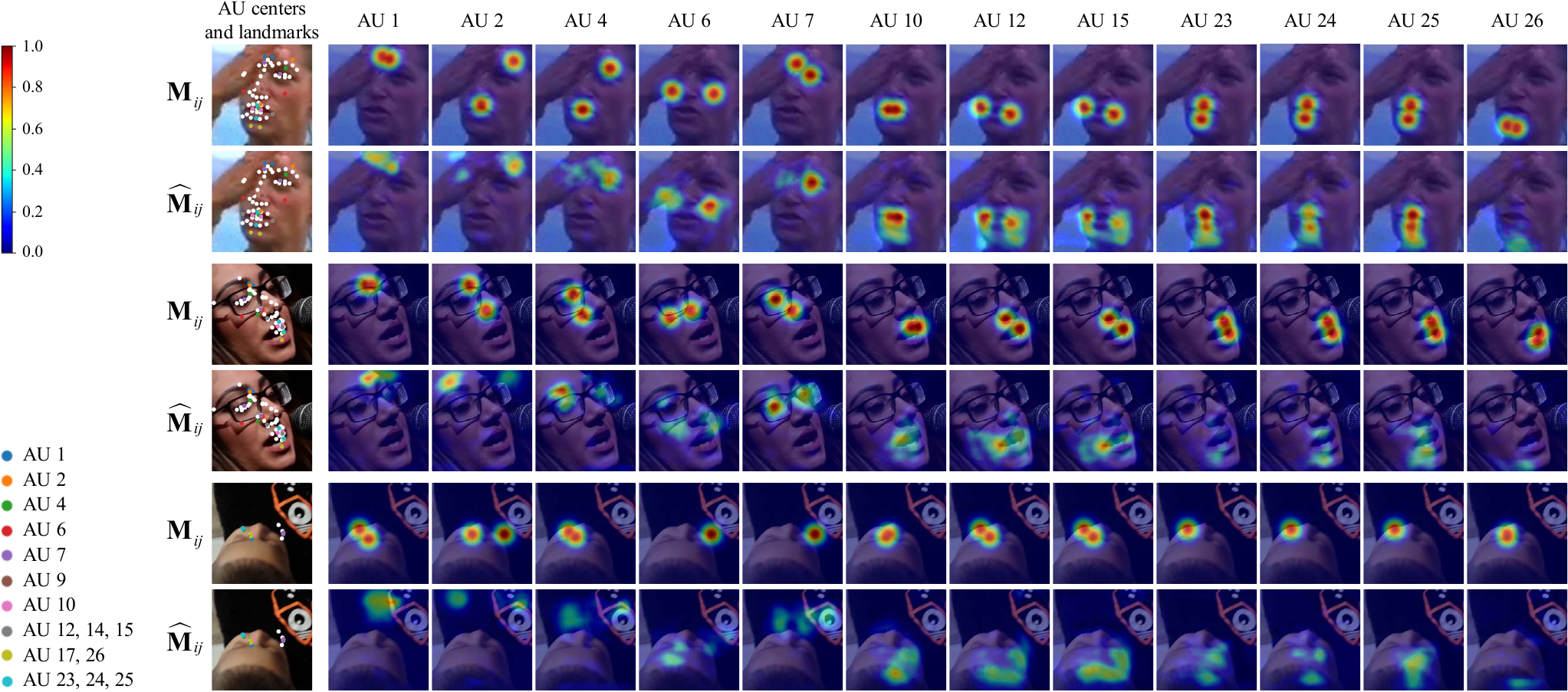}
\caption{\hhighlight{Visualization of predefined attention maps $\mathbf{M}_{ij}$ and learned attention maps $\widehat{\mathbf{M}}_{ij}$ of our AAR for three example Aff-Wild2~\cite{kollias2019expression,kollias2021analysing} images with misalignment errors. The first and second images also have partial occlusions, and the third image suffers from extreme misalignment errors. The locations of AU centers and landmarks are drawn on the images to illustrate the misalignment, and the attention weights are visualized with the colors in the color bar.}}
\label{fig:failure_attention_map}
\end{figure*}

\hhighlight{\subsection{Limitations}}

\hhighlight{
We test our AAR on input frame images with misalignment errors or occlusions. The attention maps of AAR for three example images are visualized in Fig.~\ref{fig:failure_attention_map}. Due to the mistakenly annotated landmarks, these images fail to be aligned to faces with two eye centers on the same horizontal line, eyes above nose, and fixed scale. We can see that the predefined attention maps $\mathbf{M}_{ij}$ often highlight incorrect ROIs, such as AU 2 in the first example image and AU 6 in the second example image. In contrast, given the first example image with left eye occluded by a hand and the second example image occluded by eye glasses, our AAR can almost accurately capture the ROI of each AU, as illustrated in the learned attention map $\widehat{\mathbf{M}}_{ij}$. However, if input images are severely misaligned like the third example image, AAR fails to precisely capture AU ROIs. This is partially because our AAR does not explicitly process misalignment errors, such as explicitly learning rotation-invariant and scale-invariant features. We will try to improve the robustness on severe misalignment errors and occlusions in the future work.}

\section{Conclusion}

In this paper, we have proposed a novel adaptive attention and relation (AAR) framework for facial AU detection. Specifically, we have introduced an adaptive attention regression network to integrate the advantages of local attention predefinition and global attention learning, which is beneficial for capturing both \Highlight{predefined dependencies by landmarks in} strongly correlated regions and \Highlight{facial globally distributed dependencies in} weakly correlated regions. 
Moreover, we have proposed an adaptive spatio-temporal graph convolutional network to simultaneously reason the AU-specific patterns, the inter-dependencies among AUs, as well as the temporal dependencies. To our knowledge, the specific \Highlight{way} of each AU reasoned in graph neural networks has not been done before.

We have compared our method with state-of-the-art approaches on the challenging BP4D, DISFA, GFT, and Aff-Wild2 benchmarks in both constrained and unconstrained scenarios, in which our method \highlight{obtains competitive performance.} 
Moreover, we have conducted an ablation study which demonstrates that each component in our framework contributes to AU detection. Besides, the visual results indicate that our method can accurately learn the regional correlation distribution of each AU.

\highlight{Considering the powerful sequence modeling ability of the prevailing transformer~\cite{vaswani_attention_2017},
in the future work we will explore the reasoning of temporal dependencies by transformers. We can implement our proposed idea by designing a new spatio-temporal transformer to simultaneously reason the specific pattern of each AU, the inter-dependencies among AUs, as well as the temporal dependencies.}


%




\ifCLASSOPTIONcaptionsoff
  \newpage
\fi



\bibliographystyle{IEEEtran}
\bibliography{IEEEabrv,references}
%



%

\begin{IEEEbiography}[{\includegraphics[width=1in,height=1.25in,clip,keepaspectratio]{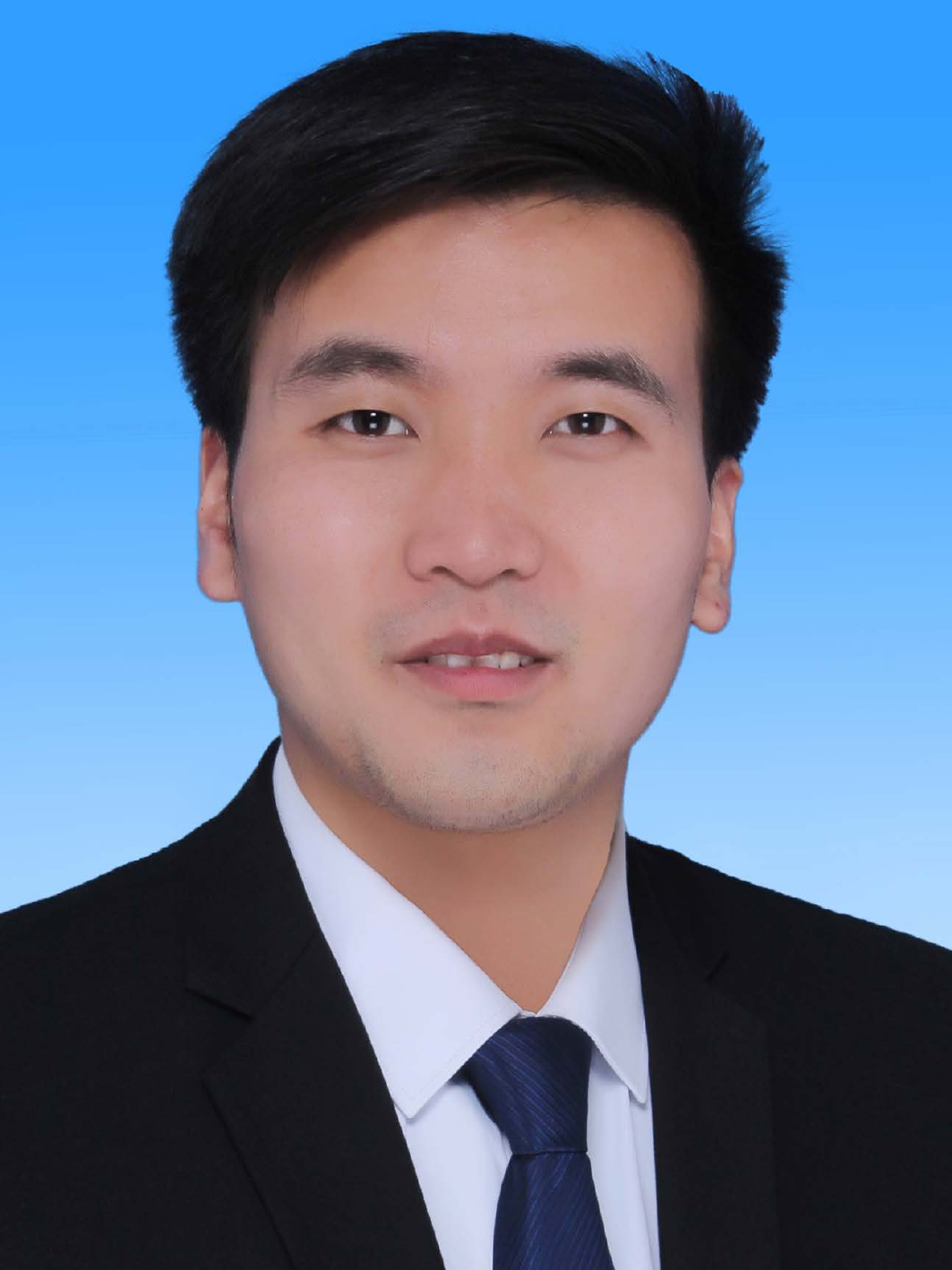}}]{Zhiwen Shao}
received his B.Eng. degree in Computer Science and Technology from the Northwestern Polytechnical University, China in 2015. He received the Ph.D. degree from the Shanghai Jiao Tong University, China in 2020. He is now an Associate Professor at the School of Computer Science and Technology, China University of Mining and Technology, China, as well as a Postdoctoral Fellow at the Department of Computer Science and Engineering, Shanghai Jiao Tong University, China. From 2017 to 2018, he was a joint Ph.D. student at the Multimedia and Interactive Computing Lab, Nanyang Technological University, Singapore. He has published more than 30 academic papers in popular journals and conferences.
His research interests lie in face analysis and deep learning. 
He has been serving as a PC member in IJCAI and AAAI.
\end{IEEEbiography}

\begin{IEEEbiography}[{\includegraphics[width=1in,height=1.25in,clip,keepaspectratio]{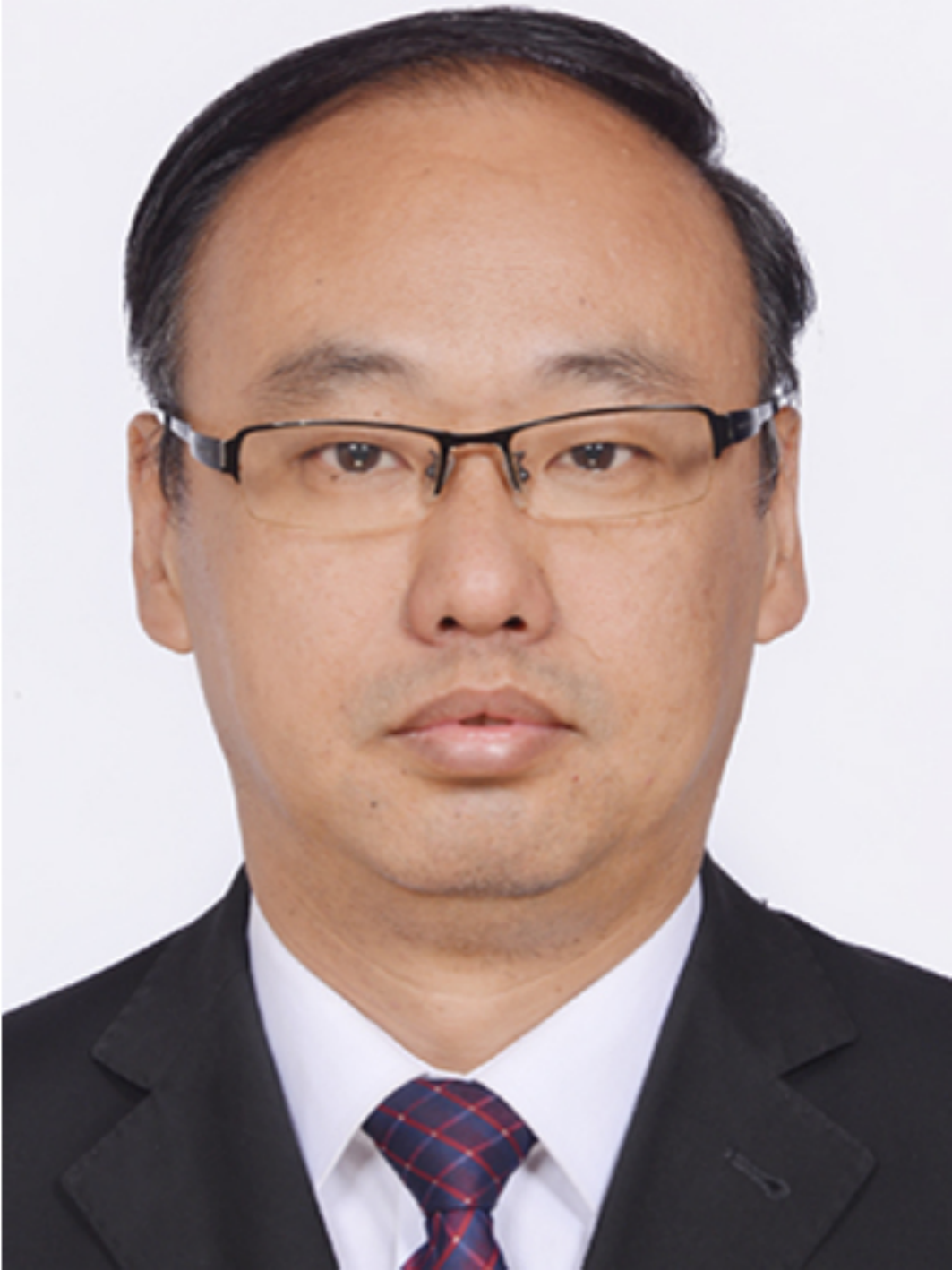}}]{Yong Zhou} received the M.S. and Ph.D. degrees in Control Theory and Control Engineering from the China University of Mining and Technology, China in 2003 and 2006, respectively. He is currently a Professor with the School of Computer Science and Technology, China University of Mining and Technology, China. His research interests include machine learning, intelligence
optimization, and data mining.
\end{IEEEbiography}

\begin{IEEEbiography}[{\includegraphics[width=1in,height=1.25in,clip,keepaspectratio]{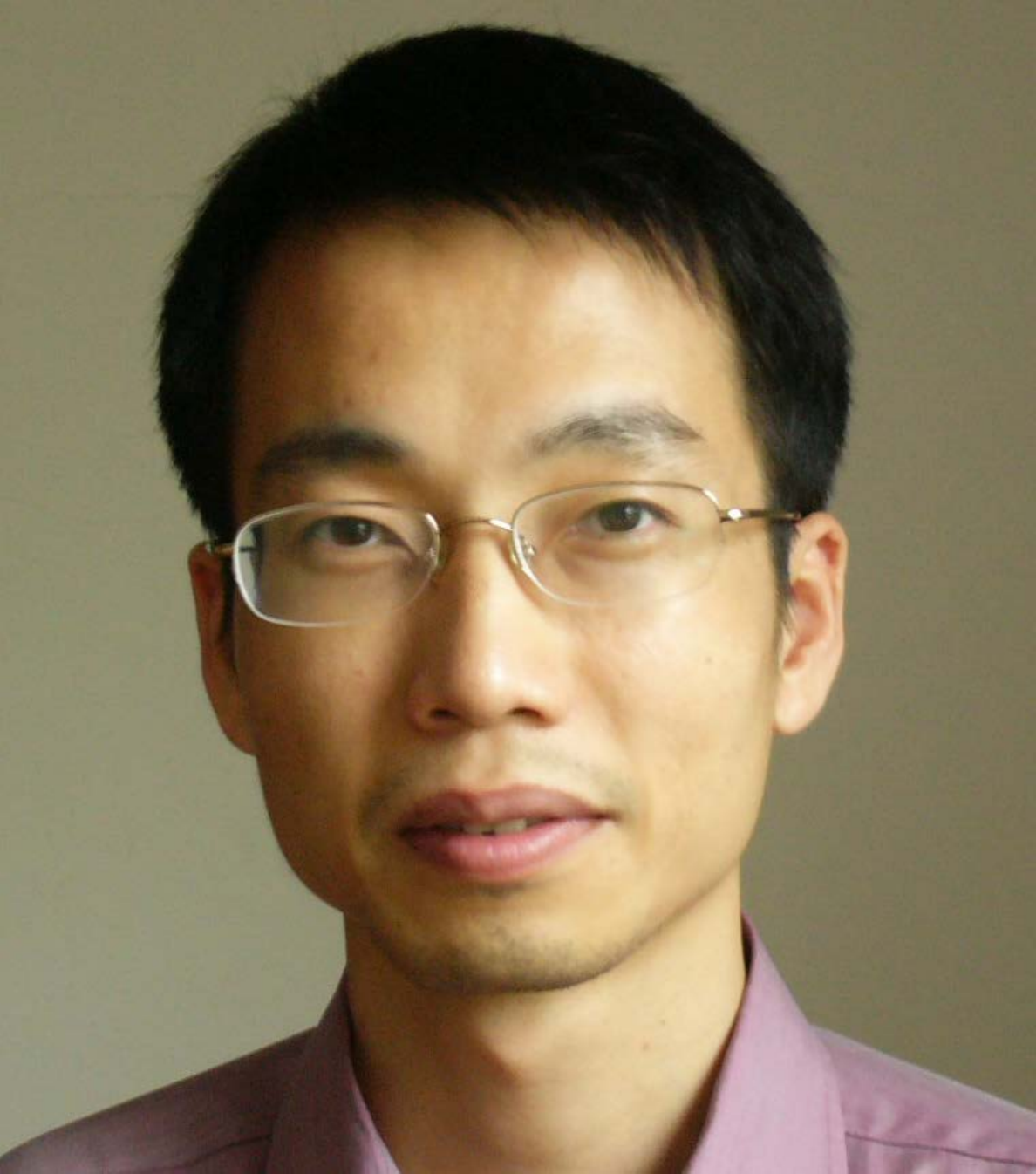}}]{Jianfei Cai}
received his Ph.D. degree from the University of Missouri-Columbia in 2002. He is currently a Full Professor and has served as the Group Lead for the Data Science \& Artificial Intelligence Group at the Faculty of Information Technology, Monash University, Australia. Before that, he was a Full Professor and a Cluster Deputy Director of Data Science \& Artificial Intelligence Research Centre at the Nanyang Technological University, Singapore. He has published over 200 technical papers in international journals and conferences. His major research interests include computer vision, multimedia and deep learning. He is an IEEE Fellow. He has been serving as Associate Editor for IEEE T-IP, T-MM, and T-CSVT as well as serving as Area Chair for ICCV, ECCV, ACM Multimedia, ICME and ICIP.
\end{IEEEbiography}

\begin{IEEEbiography}[{\includegraphics[width=1in,height=1.25in,clip,keepaspectratio]{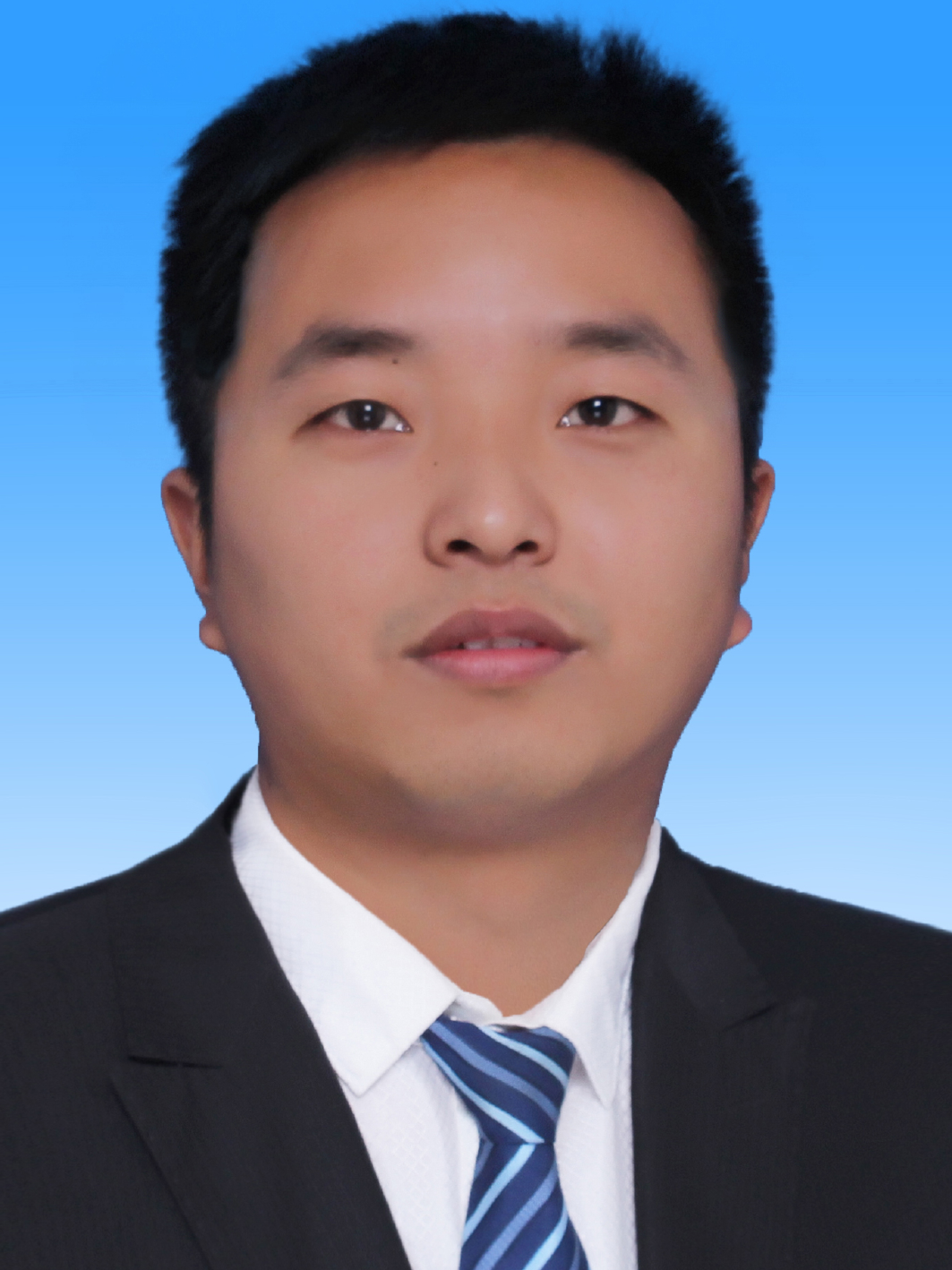}}]{Hancheng Zhu} received the B.S. degree from the Changzhou Institute of Technology, Changzhou, China, in 2012, and the M.S. and Ph.D. degrees from the China University of Mining and Technology, China, in 2015 and 2020, respectively. He is currently a Postdoctoral Fellow at the School of Computer Science and Technology, China University of Mining and Technology, China. His research interests include image aesthetics assessment and affective computing.
\end{IEEEbiography}

\begin{IEEEbiography}[{\includegraphics[width=1in,height=1.25in,clip,keepaspectratio]{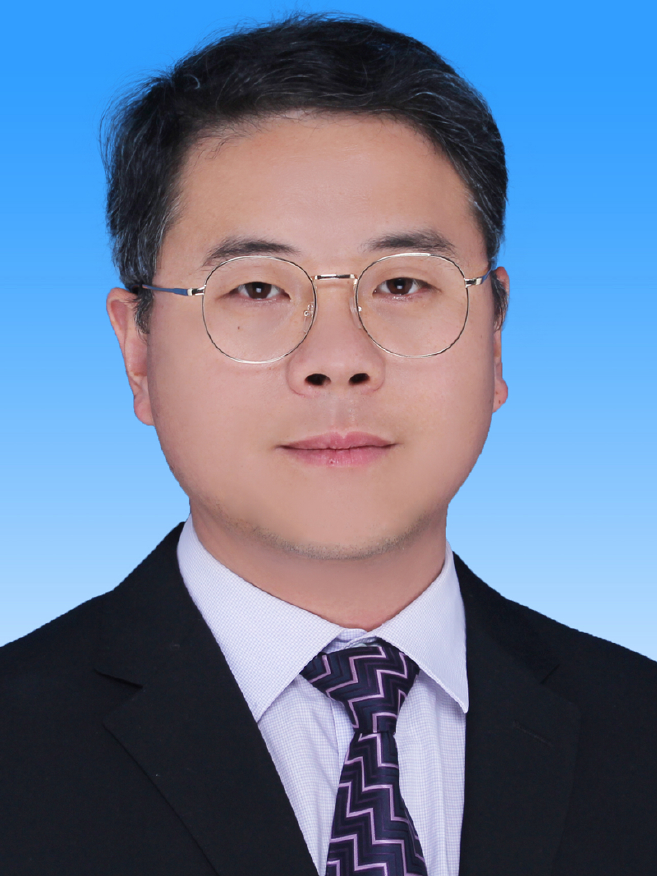}}]{Rui Yao} received the Ph.D. degree in computer science from the Northwestern Polytechnical University, Xi'an, China, in 2013. From 2011 to 2012, he was a Visiting Student with the University of Adelaide, Adelaide, SA, Australia. He is currently a Professor with the School of Computer Science and Technology, China University of Mining and Technology, China. His research interests include computer vision and machine learning.
\end{IEEEbiography}








\end{document}